\documentclass[acmtog]{acmart}
\pdfoutput=1
\usepackage{url}
\usepackage{multirow}
\usepackage{listings}
\usepackage[ruled]{algorithm2e}
\usepackage{makecell}
\usepackage{pifont}
\usepackage{color}
\usepackage{hyperref}

\AtBeginDocument{%
  \providecommand\BibTeX{{%
    \normalfont B\kern-0.5em{\scshape i\kern-0.25em b}\kern-0.8em\TeX}}}

\setcopyright{acmcopyright}
\copyrightyear{2021}
\acmYear{2021}
\acmDOI{10.1145/3474088}

\acmJournal{TOG}
\acmVolume{37}
\acmNumber{4}
\acmArticle{15}
\acmMonth{6}

\acmSubmissionID{TOG-20-0085}

\begin{document}

\title{Optical aberrations Correction in Postprocessing using Imaging Simulation}

\author{Shiqi Chen}
\orcid{0000-0001-6396-1595}
\email{chenshiqi@zju.edu.cn}
\affiliation{%
  \institution{Zhejiang University}
  \department{State Key Laboratory of Modern Optical Instrumentation}
  \city{Hangzhou}
  \state{Zhejiang}
  \country{China}
}

\author{Huajun Feng}
\email{fenghj@zju.edu.cn}
\affiliation{%
  \institution{Zhejiang University}
  \department{State Key Laboratory of Modern Optical Instrumentation}
  \city{Hangzhou}
  \state{Zhejiang}
  \country{China}
}

\author{Dexin Pan}
\email{pdx@zju.edu.cn}
\affiliation{%
  \institution{Zhejiang University}
  \department{State Key Laboratory of Modern Optical Instrumentation}
  \city{Hangzhou}
  \state{Zhejiang}
  \country{China}
}

\author{Zhihai Xu}
\email{xuzhh@zju.edu.cn}
\affiliation{%
  \institution{Zhejiang University}
  \department{State Key Laboratory of Modern Optical Instrumentation}
  \city{Hangzhou}
  \state{Zhejiang}
  \country{China}
}

\author{Qi Li}
\email{liqi@zju.edu.cn}
\affiliation{%
  \institution{Zhejiang University}
  \department{State Key Laboratory of Modern Optical Instrumentation}
  \city{Hangzhou}
  \state{Zhejiang}
  \country{China}
}

\author{Yueting Chen}
\orcid{0000-0002-2759-9784}
\email{chenyt@zju.edu.cn}
\affiliation{%
  \institution{Zhejiang University}
  \department{State Key Laboratory of Modern Optical Instrumentation}
  \city{Hangzhou}
  \state{Zhejiang}
  \country{China}
}

\authorsaddresses{%
Authors’ addresses: S. chen, H. Feng, D. Pan, Z. Xu, Q. Li, and Y. Chen, Zhejiang University; emails: chenshiqi@zju.edu.cn, fenghj@zju.edu.cn, pdx@zju.edu.cn, xuzh@zju.edu.cn, liqi@zju.edu.cn, chenyt@zju.edu.cn}
\renewcommand{\shortauthors}{Chen, et al.}
\begin{abstract}
  % influence of optical aberrations
  As the popularity of mobile photography continues to grow, considerable effort is being invested in the reconstruction of degraded images. Due to the spatial variation in optical aberrations, which cannot be avoided during the lens design process, recent commercial cameras have shifted some of these correction tasks from optical design to postprocessing systems. However, without engaging with the optical parameters, these systems only achieve limited correction for aberrations.
  
  % our method
  In this work, we propose a practical method for recovering the degradation caused by optical aberrations. Specifically, we establish an imaging simulation system based on our proposed optical point spread function (PSF) model. Given the optical parameters of the camera, it generates the imaging results of these specific devices. To perform the restoration, we design a spatial-adaptive network model on synthetic data pairs generated by the imaging simulation system, eliminating the overhead of capturing training data by a large amount of shooting and registration.
  
  % experiments and results
  Moreover, we comprehensively evaluate the proposed method in simulations and experimentally with a customized digital-single-lens-reflex (DSLR) camera lens and HUAWEI HONOR 20, respectively. The experiments demonstrate that our solution successfully removes spatially variant blur and color dispersion. When compared with the state-of-the-art deblur methods, the proposed approach achieves better results with a lower computational overhead. Moreover, the reconstruction technique does not introduce artificial texture and is convenient to transfer to current commercial cameras. Project Page: \url{https://github.com/TanGeeGo/ImagingSimulation}.
\end{abstract}

\begin{CCSXML}
	<ccs2012>
	<concept>
	<concept_id>10010147.10010178.10010224.10010245.10010254</concept_id>
	<concept_desc>Computing methodologies~Reconstruction</concept_desc>
	<concept_significance>500</concept_significance>
	</concept>
	</ccs2012>
\end{CCSXML}

\ccsdesc[500]{Computing methodologies~Reconstruction}

\keywords{optical aberrations, imaging simulation, deep-learning networks, image reconstruction}

\begin{teaserfigure}
  \includegraphics[width=\textwidth]{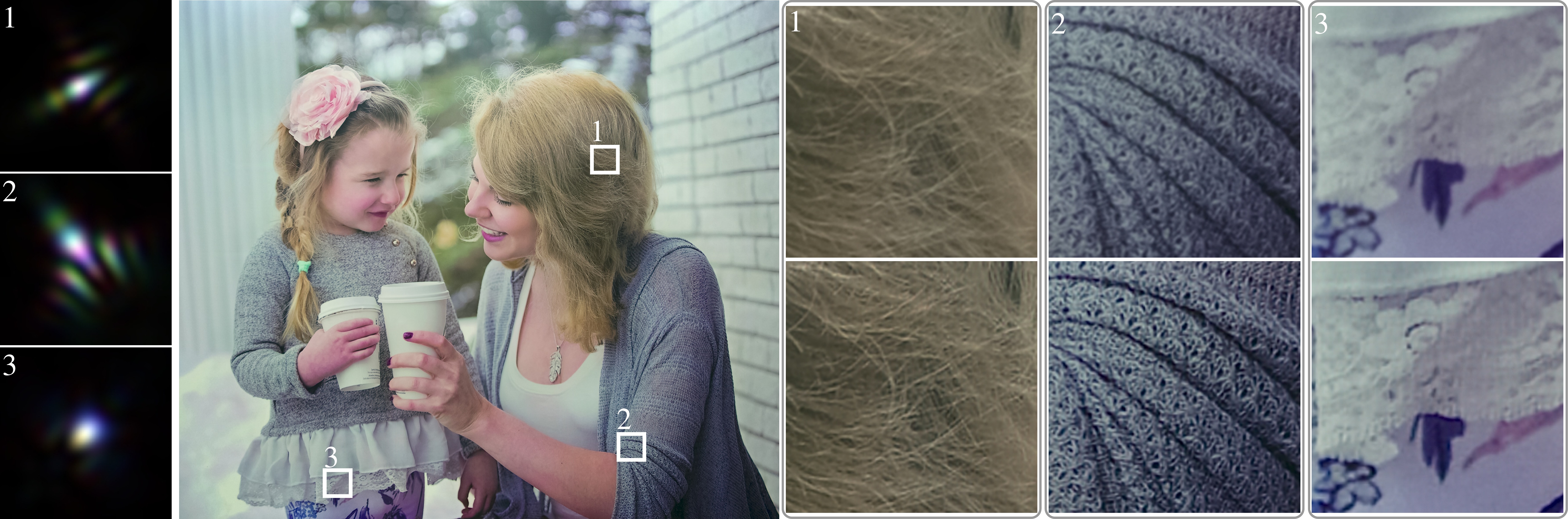}
  \caption{We present a practical method for optical aberrations correction. We calculate the spatially variant PSFs (shown on the left in $10\times$resampling) of a specific lens by the proposed optical PSF model (Section \ref{sec:Imaging Simulation System}). The PSFs are used for imaging simulation to generate training data for the proposed postprocessing chain (Section \ref{sec:aberrations Image Recovery}). Our approach significantly improves image quality (shown on the right) and is convenient for deploying to new devices.}
  \Description{\textbf{Optical Aberration Correction With Imaging Simulation.}}
  \label{fig:teaser}
\end{teaserfigure}

\maketitle

\section{Introduction}
%%%%%%%%%%%%%%%%%%%%%%%%%%%%%%%%%%%%%%%%%%%%%%%%%%%%%%%%%%%%%%%%%%%%%%%%%%%%%
%%%%%%%%%%%%%%%%%%%%%%%%%%%%%%%%%%%%%%%%%%%%%%%%%%%%%%%%%%%%%%%%%%%%%%%%%%%%%
\begin{figure}[t]
  \centering
  \includegraphics[width=\linewidth]{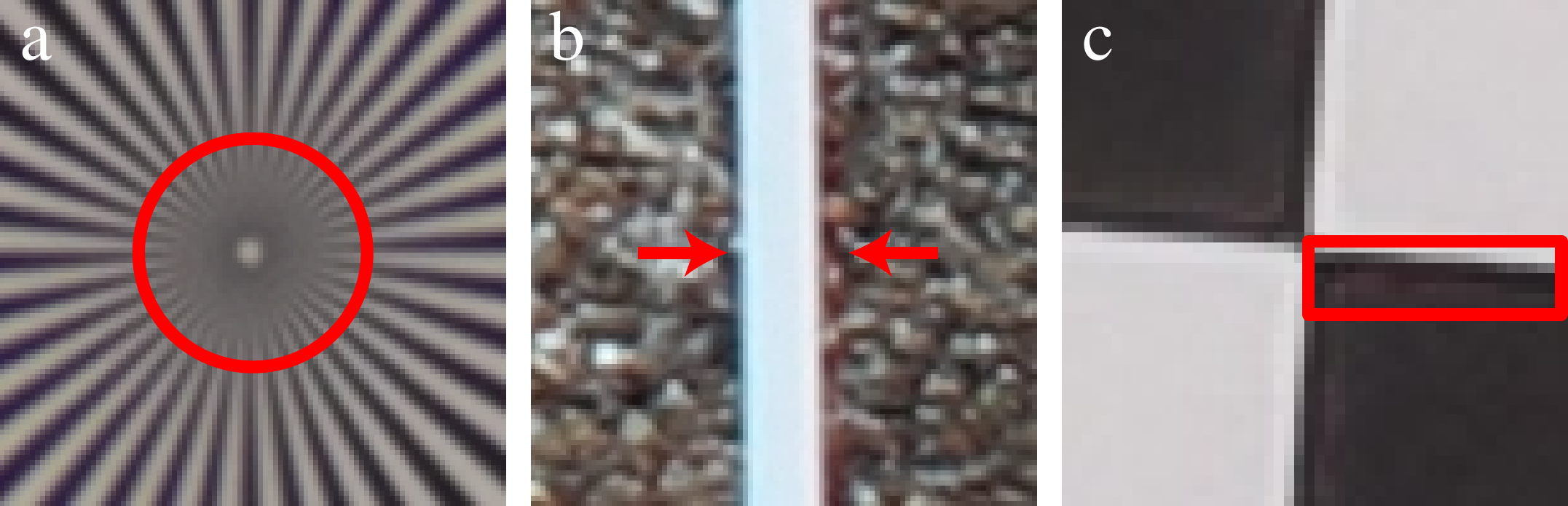}
  \caption{Typical optical aberration degradation appearing on photos. From a to c: blurring(Canon EOS 80D + EF 50mm f/1.8 STM), chromatism (Canon EOS 80D + EF 50 mm f/1.2 L USM), and "ringing effect" (HUAWEI HONOR 20). We mark typical degradation with red indicators so that readers can spot them quickly.}
  \Description{Typical optical aberration degradation appearing on photos from digital cameras.}
  \label{fig:aberration ISP}
\end{figure}
%%%%%%%%%%%%%%%%%%%%%%%%%%%%%%%%%%%%%%%%%%%%%%%%%%%%%%%%%%%%%%%%%%%%%%%%%%%%%
%%%%%%%%%%%%%%%%%%%%%%%%%%%%%%%%%%%%%%%%%%%%%%%%%%%%%%%%%%%%%%%%%%%%%%%%%%%%%
As an extension of human vision, modern digital imaging techniques have greatly enriched our methods of interacting with the world. Coupled with complex optical systems, powerful digital sensors, and heavy image signal processing (ISP) algorithms, today's digital cameras can handle many extreme shooting scenarios, such as low-light \cite{chen2018learning}, long-distance \cite{zhang2019zoom} and high-dynamic-range scenarios \cite{yan2019attention,hasinoff2016burst}. However, sophisticated as today's digital imaging systems are, it is difficult to totally avoid the degradation caused by optical aberration. To alleviate the aberrations while satisfying the constraints of spacing, mobile imaging devices (smartphones, DSLR cameras, {\itshape etc.}) adopt many different approaches. For example, the professional large aperture lens replaces ordinary glass with fluorite, and the optical stack of HUAWEI HONOR 20 has more than eight extended aspherical elements, whose aspheric coefficients are up to order 16 \cite{zhenggang2016periscope}.

In addition to the optical lens module, ISP algorithms also make a large difference in optical aberration correction. However, as shown in Figure \ref{fig:aberration ISP}a and Figure \ref{fig:aberration ISP}b, the blur and chromatic dispersion vary greatly when the optical system changes. Moreover, the current classical ISP systems still cannot handle these issues completely, therefore introducing the "ringing effect" (see Figure \ref{fig:aberration ISP}c). These defects are very common in high-contrast photos from many recent flagship devices.

Over the past decade, methods based on the deep learning model have been evolving rapidly and have the potential to replace conventional ISP systems \cite{ignatov2020replacing}. The classical postprocessing system is a step-by-step process. The optical aberrations of lens and the error of noise diagnosis will accumulate and be amplified in the subsequent white balance (WB), color correction matrix (CCM), gamma operations, and so on. \cite{heide2014flexisp}. Nevertheless, end-to-end deep learning methods do not suffer from this problem \cite{2018A}. Although deep learning networks allow us to restore latent images from degraded images, the learned models always achieve unsatisfactory results when evaluating real data \cite{2020Toward}. Therefore, generating targeted datasets that follow the distribution of real scenes is the principal part of the deep learning method \cite{2020Real}. In recent years, many scholars have gained insight into this field. Their proposed approaches are characterized by a considerable amount of shooting and alignment \cite{ignatov2017dslr}. However, a large quantity of training data is required to prevent overfitting, which conversely exacerbates the workload \cite{peng2019learned}. Furthermore, because each imaging system exhibits entirely different characteristics, adapting a CNN-based algorithm to a new camera may require capturing many new data pairs \cite{ignatov2019ntire}. Therefore, it is imperative to design a brand new method for dataset generation that combines optical parameters and is easy to transfer to other cameras.

In this work, we propose a practical method for recovering the degradation caused by optical aberrations. To achieve this, we establish an imaging simulation system based on raytracing and coherent superposition \cite{paul2014huygens}. Given the lens data and parameters of the sensor, it works out the imaging result of a definite object at a fixed distance. With datasets established by the imaging simulation system, we design a postprocessing chain, which combines field-of-view (FOV) information and optical parameters to perform adaptive restoration on the degraded data \cite{heide2014flexisp}. The proposed method outperforms many state-of-the-art deblurring methods on synthetic data pairs, as does the evaluation. When evaluated on real images, our proposed technique achieves better visual results and requires lower computational overhead. Moreover, our approach is convenient for transferring to other commercial lenses, and there are no limitations on the optical design.

Our main technical contributions are as follows:

\begin{itemize}
  \item An optical point spread function model is proposed for accurate PSF calculation in a simulation manner.

  \item Based on the ISP algorithms, an imaging simulation system is established. Given the camera's optical parameters, it generates targeted data pairs following the distribution of lens optical aberrations.
  
  \item A novel spatial-adaptive CNN architecture is proposed, which introduces FOV input, deformable ResBlock, and context block to adapt to spatially variant degradation.
\end{itemize}

This paper proceeds as follows. In Section \ref{sec:Related Work}, we review the related works. In Section \ref{sec:Overview}, we present an overview of our methods. In Section \ref{sec:Imaging Simulation System}, we detail the proposed optical PSF model and the composition of the imaging simulation system. In Section \ref{sec:aberrations Image Recovery} and Section \ref{sec:Data Preparation}, we present a brand new postprocessing pipeline. In Section \ref{sec:Analysis}, we compare our approach's improved performance with the state-of-the-art methods and perform an ablation study on each aspect of our approach. In Section \ref{sec:Experimental Assessment}, we assess the proposed approach and demonstrate its potential applications.
%%%%%%%%%%%%%%%%%%%%%%%%%%%%%%%%%%%%%%%%%%%%%%%%%%%%%%%%%%%%%%%%%%%%%%%%%%%%%
%%%%%%%%%%%%%%%%%%%%%%%%%%%%%%%%%%%%%%%%%%%%%%%%%%%%%%%%%%%%%%%%%%%%%%%%%%%%%
\section{Related Work}
\label{sec:Related Work}

{\itshape Optical Aberration.} According to Fermat's principle, rays of light traverse the path of stationary optical path length with respect to variations in the path \cite{westphal2002correction}. Therefore, when the incident angle of rays changes, so does the optical length \cite{smith2005modern}. Generally, the distinction of optical length manifests as weird blur and pixel displacement, which becomes more severe with increasing numerical aperture and FOV \cite{smith2008modern}. While limited by the requirements of physical space, optical parameters, and machining tolerance, it is difficult for classical lens design to suppress optical aberrations \cite{sliusarev1984aberration}.

{\itshape Point Spread Function Model.} A large body of work in this field has proposed calibrating and measuring PSFs from a single image \cite{hirsch2015self,shih2012image, jemec20172d,ahi2017mathematical}. These methods often estimate optical blur functions or chromatic aberrations either blindly or through a calibration process. Moreover, recent works propose developing a machine learning framework to estimate the underlying physical PSF model from the given camera lens \cite{herbel2018fast}. However, these approaches have several limitations in operation, such as the noise level and the field interval of calibration or estimation \cite{heide2013high}. The closest related works to us are those approaches that compute PSFs with lens prescriptions \cite{shih2012image}. Given the focal length, aperture, focusing distance, and white balance information, these methods simulate the image plane PSFs of each virtual object point.

Unfortunately, the PSFs calculated by their method are different from the measured PSFs (after ISP). The reasons for mismatching are summarized as manufacturing tolerances and fabrication errors during lens production \cite{shih2012image}. However, we hold different views from them. We believe that the PSFs calculated by lens prescription represent the diffusion of energy, which are essentially different from the measured PSFs after ISP. \cite{paulin2008point}. And the complex amplitude of each ray must be considered while calculating the PSFs. Therefore, coherent superposition must be conducted after raytracing. To make a comprehensive comparison to prove the accuracy of the computed energy PSFs, we apply the calculated PSFs to the checkerboard chart in the energy domain and then transform the degraded image to a visualized image. The similarity of the real images and the synthetic results illustrates that it is a practical method for simulating PSFs with the advantages of lens prescriptions if the process of imaging is properly modeled.

{\itshape Image Reconstruction.} The optical aberration correction of a real image can be regarded as a problem of image reconstruction, where the goals are to increase sharpness and restore detailed features of images. Traditional deconvolution methods \cite{heide2013high,pan2016blind,sun2017revisiting} use various natural image priors for iterative and mutual optimization, which unfortunately are not robust and have poor computational efficiency when handling large spatially variant blur kernels. The point spread functions of large FOV lenses vary from 11 to 53 pixels, which presents a great challenge to existing deconvolution methods, necessitating a custom image reconstruction approach. In recent years, a large number of efficient solutions have been proposed to use deep learning models for image processing tasks \cite{schuler2015learning,agustsson2017ntire,nah2017deep,vedaldi2010vlfeat}, starting from the simplest CNN approaches \cite{sun2015learning,dong2015image,kim2016accurate,shi2016real}, to deep residual and attention models \cite{zhang2019zoom,wang2019edvr,ratnasingam2019deep} and complex generative adversarial networks (GANs) \cite{ledig2017photo,sajjadi2017enhancenet,wang2018esrgan,ignatov2020replacing}. More recently, \cite{peng2019learned} proposed a learned generative reconstruction model, a single deep Fresnel lens design tailored to this model, obtaining pleasant results.

All of these approaches have in common that they require either accurate PSF estimation \cite{heide2013high} or large training data that have been manually acquired \cite{ignatov2020replacing}. Moreover, when the lens design changes, painful acquisition is still needed to obtain targeted image pairs, which include shooting, registration, and color correction \cite{peng2019learned}. In contrast, our proposed imaging simulation system generates a large number of training corpora with encoded optical information. Moreover, our approach is practical and easy to transfer to other lens designs. Our network model combines the optical parameters and FOV information, allowing us to efficiently address the large scene-dependent blur and color shift of commercial lens design.

{\itshape Mobile Image Signal Processing.} There exist many classical approaches for various image signal processing subtasks, such as image demosaicing \cite{hirakawa2005adaptive,dubois2006filter,li2008image}, denoising \cite{buades2005non,foi2008practical,condat2010simple,chang2020spatial}, white balancing \cite{schwartzburg2014high,van2007edge,gijsenij2011improving}, and color correction \cite{kwok2013simultaneous}. Only a few works have explored the applicability of combining optical parameters with imaging signal processing systems. Recently, some deep learning methods were proposed to map low-quality smartphone RGB/RAW photos to superior-quality images obtained with a high-end reflex camera \cite{ignatov2017dslr}. The collected datasets were later used in many subsequent works \cite{ignatov2018wespe,ignatov2020replacing, mei2019higher}, which achieved significantly improved results. Additionally, in \cite{agustsson2017ntire,ignatov2019ntire}, the authors examined the possibility of running image enhancement models directly on smartphones and proposed several efficient solutions for this task.

It is worth mentioning that all these datasets' construction methods require a considerable amount of shooting and registration. Moreover, owing to the different perspectives of data pairs, the resulting image enhancement model will introduce a small displacement to the input data, which is different from lens distortion correction. Additionally, these methods map the degradation images shot by a poor imaging system to the photos shot by a better imaging system; in other words, the upper bound of recovery depends on the quality of the ground truth. In contrast, the quality of the ground truth is also important but not critical to our method, because we encode the optical aberration degradation in the PSFs instead of the image pairs. And our approach generates data in a simulation manner, which liberates us from many duplicate acquisitions.
%%%%%%%%%%%%%%%%%%%%%%%%%%%%%%%%%%%%%%%%%%%%%%%%%%%%%%%%%%%%%%%%%%%%%%%%%%%%%
%%%%%%%%%%%%%%%%%%%%%%%%%%%%%%%%%%%%%%%%%%%%%%%%%%%%%%%%%%%%%%%%%%%%%%%%%%%%%
\section{Overview}
\label{sec:Overview}
%%%%%%%%%%%%%%%%%%%%%%%%%%%%%%%%%%%%%%%%%%%%%%%%%%%%%%%%%%%%%%%%%%%%%%%%%%%%%
%%%%%%%%%%%%%%%%%%%%%%%%%%%%%%%%%%%%%%%%%%%%%%%%%%%%%%%%%%%%%%%%%%%%%%%%%%%%%
\begin{figure*}[t]
  \centering
  \includegraphics[width=\linewidth]{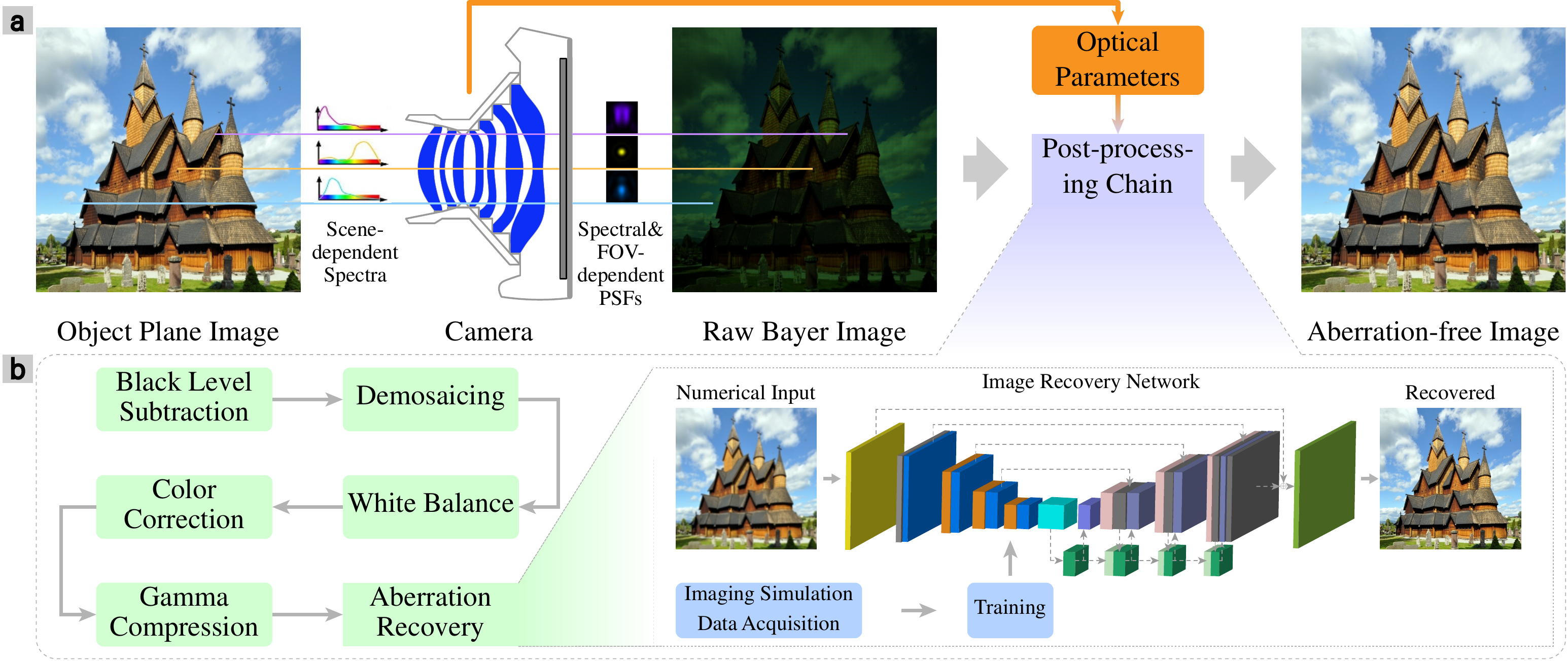}
  \caption{\textbf{Overview.} The image formation model and the postprocessing chain are sketched in (\textbf{a}). In the image formation process, the camera lens encodes spectrally and field-of-view dependent PSFs in the resulting raw Bayer image. In the postprocessing chain, the optical parameters of the lens are applied to decode the spatially variant blurring and chromatic aberrations. The proposed postprocessing chain is detailed in (\textbf{b}). It jointly combines synthetic data acquisition, spatial-adaptive network architecture and optical parameters to recover latent images.}
  \Description{Overview.}
  \label{fig:Overview}
\end{figure*}
%%%%%%%%%%%%%%%%%%%%%%%%%%%%%%%%%%%%%%%%%%%%%%%%%%%%%%%%%%%%%%%%%%%%%%%%%%%%%
%%%%%%%%%%%%%%%%%%%%%%%%%%%%%%%%%%%%%%%%%%%%%%%%%%%%%%%%%%%%%%%%%%%%%%%%%%%%%

The goal of our work is to propose a practical method for recovering the degradation caused by optical aberrations. To achieve this, we engage optical parameters with a brand new postprocessing chain (as shown in Figure \ref{fig:Overview}.b). The two core ideas behind the proposed method are as follows: first, to eliminate the blurring, displacement, and chromatic aberrations caused by lens design, we design a spatially adaptive network architecture and insert it into the postprocessing chain. Second, for the sake of portability, we design an imaging simulation system for dataset generation, which simulates the imaging results of a specific camera and generates a large training corpus for deep learning methods.

Therefore, we rely on the proposed optical PSF model to establish the imaging simulation system. Different from the traditional raytracing or fast Fourier transform (FFT) methods to calculate PSF, we regard each ray as a secondary light source. The complex amplitude of each ray is superposed on the image plane; therefore, the PSF is the intensity of complex amplitude after superposition. In this way, the proposed method determines the degenerative consequences of object points. Figure \ref{fig:PSF characteristic} shows the spatially variant point spread functions of the customized DSLR camera lens and HUAWEI HONOR 20, which are computed following our optical PSF model. After calculating the PSFs of the lens, we establish an imaging simulation system based on the ISP pipeline of the digital camera. The simulation pipeline is detailed in Figure \ref{fig:Imaging Simulation System}. To the best of our knowledge, we are the first to generate datasets of optical aberrations in a simulation manner, which do not need a great deal of shooting, registration, and color correction. We emphasize that our simulation method is easy to transfer to other lenses and has no limitations on optical designs.

Because of the spatially variant characteristics of optical aberrations, we design a brand new spatial-adaptive network architecture that takes the degraded image and FOV information as the input to directly reconstruct the aberration-free image. Together with the datasets generated by the imaging simulation system and the newly proposed postprocessing chain (Figure \ref{fig:Overview}.b), we can address the spectral and FOV-dependent blurring, color shift, and sharpness loss caused by lens aberrations.

%%%%%%%%%%%%%%%%%%%%%%%%%%%%%%%%%%%%%%%%%%%%%%%%%%%%%%%%%%%%%%%%%%%%%%%%%%%%%
%%%%%%%%%%%%%%%%%%%%%%%%%%%%%%%%%%%%%%%%%%%%%%%%%%%%%%%%%%%%%%%%%%%%%%%%%%%%%
%%%%%%%%%%%%%%%%%%%%%%%%%%%%%%%%%%%%%%%%%%%%%%%%%%%%%%%%%%%%%%%%%%%%%%%%%%%%%
\section{Imaging Simulation System}
\label{sec:Imaging Simulation System}

In this section, we start by introducing the proposed optical PSF model, which consists of raytracing and coherent superposition. Second, we illustrate the relative illumination distribution and sensor spectral response calculation, which are used to connect the separate PSFs of different wavelengths to the accurate energy PSFs. Third, based on the ISP systems of modern digital cameras, we design an imaging simulation pipeline (Figure \ref{fig:Imaging Simulation System}), which is applied to the input images, and generate the results degraded by PSFs depending on the spectral distribution and FOV information.

\subsection{Optical PSF Model}\label{sec:Optical PSF Model}
For photographic objective systems with large FOVs and apertures, it is inaccurate to derive the PSF analytic solution by Gaussian paraxial approximation \cite{born2013principles} or fast Fourier transform (FFT) \cite{manual2009optical} (\textbf{we recommend readers to supplementary materials for relevant information}). In other words, for various commercial optical designs, sequential raytracing and coherent superposition are the only universal methods to obtain precise point spread functions on the imaging plane.

In the image formation process, every point of the object can be regarded as a coherent light source $\mathcal{S}$. For a particular light source $\mathcal{S}$, a grid of rays is launched through the optical system, which forms the wavefront $\mathcal{\sum}$ at the exit pupil. According to the Huygens principle, every point on a wavefront is the source of spherical wavelets, and the secondary wavelets emanating from different points mutually interfere \cite{baker2003mathematical, paul2014huygens}. In other words, the diffraction intensity at any point on the image plane is the complex sum of all these wavelets, squared. Therefore, in our proposed optical PSF model, the PSF on the image grid is computed in two stages, the first stage uses the sequential raytracing method to compute the wavefront distribution at the exit pupil plane, and the second stage uses coherent superposition to output the PSF calculation. 
%%%%%%%%%%%%%%%%%%%%%%%%%%%%%%%%%%%%%%%%%%%%%%%%%%%%%%%%%%%%%%%%%%%%%%%%%%%%%
%%%%%%%%%%%%%%%%%%%%%%%%%%%%%%%%%%%%%%%%%%%%%%%%%%%%%%%%%%%%%%%%%%%%%%%%%%%%%
\subsubsection{Raytracing}\label{sec:Ray tracing}
The skew ray emitted by a coherent monochromatic light source $\mathcal{S}$ is a perfectly general ray, which can be defined by the starting point \textbf{O} = ($x_{0}$, $y_{0}$, $z_{0}$) of $\mathcal{S}$, and by its normalized direction vector \textbf{D} = ($X$, $Y$, $Z$). We can express the ray \textbf{P} = ($x$, $y$, $z$) in parametric form, where $t$ is the ray marching distance:
\begin{equation}\label{eqt:parametric form of general light}
    \textbf{P} = \textbf{O} + t\textbf{D}, t\geq0,
\end{equation}
After defining a general ray, raytracing can be divided into two steps:
The first step is to calculate the intersection point of light and the surface. The surface equation generally used in optical design can be formulated as follows:

\begin{equation}\label{eqt:equation of aspheric}
    z = \mathit{f}(x, y) = \underbrace{\frac{c\rho^{2}}{1 + \sqrt{1-c^{2}\rho^{2}}}}_{sphere\;part} + \underbrace{A_{2}\rho^{2} + A_{4}\rho^{4} + \cdots + A_{j}\rho^{j}}_{deformation\;terms},
\end{equation}
where $z$ is the longitudinal coordinate of a point on the surface, and $\rho = \sqrt{x^{2} + y^{2}}$ is the distance from the point to $z$ axis. $c$ is the curvature of the spherical part. The subsequent terms represent deformations to the spherical part, with $A_{2}$, $A_{4}$, $etc.$, as the constant of the second, fourth, $etc$., power deformation terms.

We build the simultaneous equations of Eq. \ref{eqt:parametric form of general light} and Eq. \ref{eqt:equation of aspheric} for solving the ray marching distance from the starting point $\textbf{O}$ to the next ray-surface intersection point. However, their solution cannot be directly determined by the higher-order equations. Therefore, our raytracing is accomplished by a series of approximations. In each approximation, we add a small increment to the ray marching distance and compute \textbf{P} = ($x$, $y$, $z$) with Eq.\ref{eqt:parametric form of general light} to obtain the coordinates of ray. Additionally, we substitute $x$ and $y$ into Eq. \ref{eqt:equation of aspheric} to acquire the $z_{ref}$, which is the longitudinal coordinate of the surface. The iteration is repeated until the disparity between $z$ and $z_{ref}$ is negligible. In this way, the intersection of light and the surface is computed.

The second step is to calculate the refraction of light on the medium surface. Following Snell's law, the refracted angle $\theta_{i}$ can be computed by:

\begin{equation}\label{eqt:refracted angle}
    sin^{2}\theta_{t} = (\frac{\eta_{1}}{\eta_{2}})^{2}(1-cos^{2}<\textbf{n}, \textbf{D}>),
\end{equation}
Here, \textbf{n} is the normal unit vector of the surface equation, and $\eta_{1}$ and $\eta_{2}$ are the refractive indexes on both sides of the surface. $cos<\cdot, \cdot>$ is the operation of computing the cosine value of two vectors. In this way, the refracted direction vector \textbf{D'} can be calculated by:
\begin{equation}\label{eqt:refracted direction vector}
    \textbf{D'} = \frac{\eta_{1}}{\eta_{2}}\textbf{D} + (\frac{\eta_{1}}{\eta_{2}}cos<\textbf{n}, \textbf{D}> - \sqrt{1-sin^{2}\theta_{t}})\textbf{n},
\end{equation}
In the implementation, we perform sequential raytracing for every sample position $(x', y')$ on the pupil plane. The direction vector $D_{x'y'}$ and the total optical path length $l_{x'y'}$ are recorded. We note that every object point can be regarded as a coherent light source $\mathcal{S}$, so the diffusion of energy is the superposition result of the coherent wavelet (as demonstrated in Section. \ref{sec:Optical PSF Model}). In the following section, we introduce the details of coherent superposition.

%%%%%%%%%%%%%%%%%%%%%%%%%%%%%%%%%%%%%%%%%%%%%%%%%%%%%%%%%%%%%%%%%%%%%%%%%%%%%
\subsubsection{coherent superposition}\label{sec:coherent superposition}
%%%%%%%%%%%%%%%%%%%%%%%%%%%%%%%%%%%%%%%%%%%%%%%%%%%%%%%%%%%%%%%%%%%%%%%%%%%%%
%%%%%%%%%%%%%%%%%%%%%%%%%%%%%%%%%%%%%%%%%%%%%%%%%%%%%%%%%%%%%%%%%%%%%%%%%%%%%
\begin{figure}[t]
  \centering
  \includegraphics[width=\linewidth]{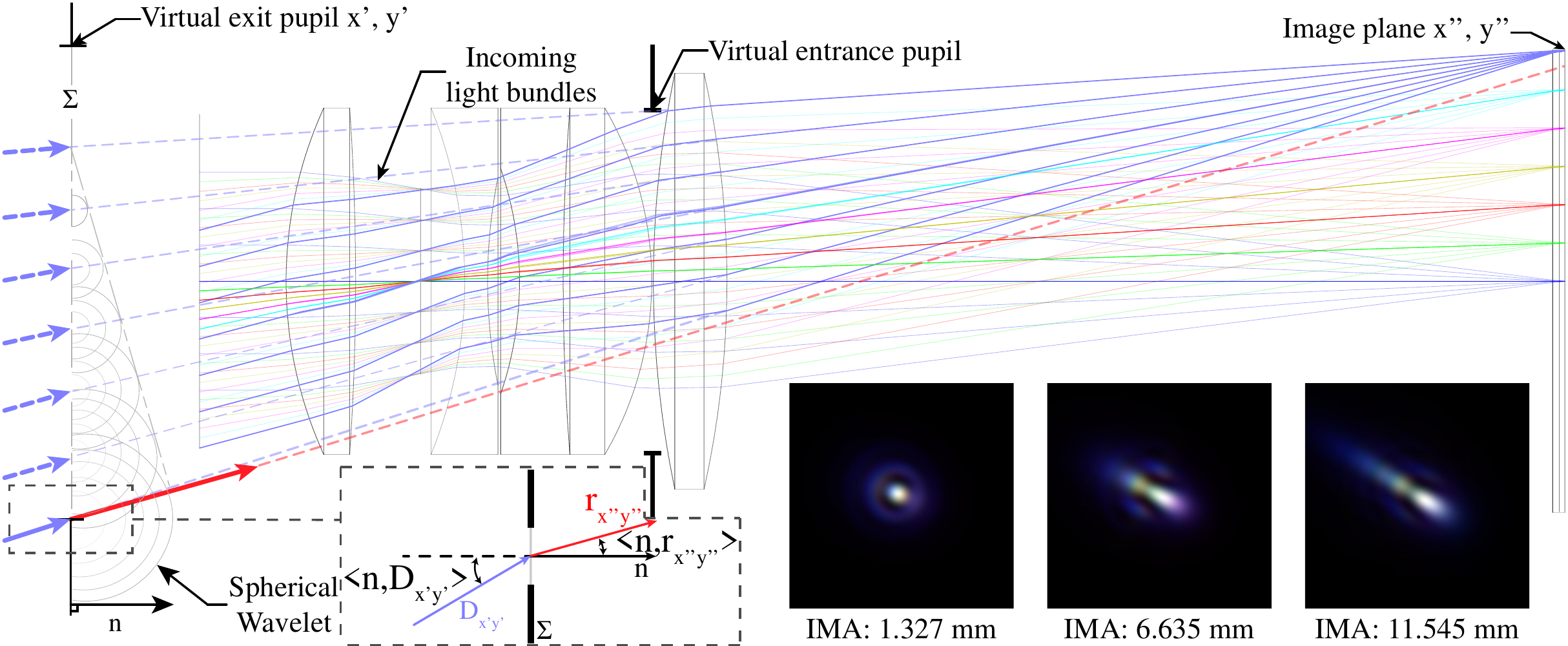}
  \caption{\textbf{Coherent superposition.} Each position uniformly sampled on wavefront distribution $\sum$ is a spherical wavelet, which coherently superposes on the image plane. In other words, the spherical wavelets interfere with each other when reaching the image plane. In the lower-right corner, we show the different fields' PSFs of customized DSLR camera lenses computed by raytracing and coherent superposition.}
  \Description{Coherent Superposition.}
  \label{fig:Coherent Superposition}
\end{figure}
%%%%%%%%%%%%%%%%%%%%%%%%%%%%%%%%%%%%%%%%%%%%%%%%%%%%%%%%%%%%%%%%%%%%%%%%%%%%%
%%%%%%%%%%%%%%%%%%%%%%%%%%%%%%%%%%%%%%%%%%%%%%%%%%%%%%%%%%%%%%%%%%%%%%%%%%%%%
%%%%%%%%%%%%%%%%%%%%%%%%%%%%%%%%%%%%%%%%%%%%%%%%%%%%%%%%%%%%%%%%%%%%%%%%%%%%%
%%%%%%%%%%%%%%%%%%%%%%%%%%%%%%%%%%%%%%%%%%%%%%%%%%%%%%%%%%%%%%%%%%%%%%%%%%%%%
\begin{algorithm}[t]
    \caption{Optical PSF model} 
    \LinesNumbered 
    \KwIn{coordinates of object point $(x_{0}, y_{0}, z_{0})$, \
    coordinates of virtual exit pupil $z'$, \
    coordinates of image plane $z''$, \
    sample range on virtual exit pupil $x'_{r} \times y'_{r}$, \
    sample interval of virtual exit pupil $\tau'$, \
    sample range on image plane $x''_{r} \times y''_{r}$, \
    sample interval of sampling range at image plane $\tau''$, \
    normal unit vector of virtual exit pupil \textbf{n}}\label{alg:PSF calculation}
    \KwOut{PSF matrix $I_{x''y''}$}
    Generate an $ceil(\frac{x''_{r}}{\tau''}) \times ceil(\frac{y''_{r}}{\tau''})$ PSF matrix $I_{x''y''}$\;
    Generate an $ceil(\frac{x''_{r}}{\tau''}) \times ceil(\frac{y''_{r}}{\tau''})$ complex amplitude matrix $\mathbf{\widetilde{E}_{x''y''}}$\;
    Compute the intersection $(x''_{c}, y''_{c}, z'')$ of the main ray and the image plane by raytracing\; 
    Initialize $\mathbf{\widetilde{E}}_{sum} = 0 + 0i$\;
    \For{$x''$ = $x''_{c}-\frac{x''_{r}}{2}$ to $x''_{c}+\frac{x''_{r}}{2}$, step: $\tau''$}{
        \For{$y''$ = $y''_{c}-\frac{y''_{r}}{2}$ to $y''_{c}+\frac{y''_{r}}{2}$, step: $\tau''$}{
            \For{$x'$ = $-\frac{x'_{r}}{2}$ to $\frac{x'_{r}}{2}$, step: $\tau'$}{
                \For{$y'$ = $-\frac{y'_{r}}{2}$ to $\frac{y'_{r}}{2}$, step: $\tau'$}{
                    Compute optical path length $\mathbf{l}_{x'y'}$ and direction cosines $\mathbf{D}_{x'y'}$ by ray tracing from point $(x_{0}, y_{0}, z_{0})$ to point $(x', y', z')$\;
                    $\mathbf{r}_{x''y''} = (x''-x', y''-y', z''-z')$\;
                    calculate $\mathbf{K}(\mathbf{D}_{x'y'}, \mathbf{r}_{x''y''}, \mathbf{n})$ using Eq.\ref{eqt:tilt factor}\;
                    calculate $\mathbf{E}(\mathbf{l}_{x'y'}, \mathbf{r}_{x''y''}, \mathbf{K})$ using Eq.\ref{eqt:complex amplitude on image plane}\;
                    $\mathbf{\widetilde{E}}_{sum} += \mathbf{E}(\mathbf{l}_{x'y'}, \mathbf{r}_{x''y''}, \mathbf{K})$\;
                }
            }
            $\mathbf{\widetilde{E}}_{x''y''} = \mathbf{\widetilde{E}}_{sum}$\;
            $\mathbf{\widetilde{E}}_{sum} = 0 + 0i$\;
        }
    }
    $I_{x''y''} = \mathbf{\widetilde{E}}_{x''y''}.*\mathbf{\widetilde{E}}^{*}_{x''y''}$\;
    \Return $I_{x''y''}$\;
\end{algorithm}
%%%%%%%%%%%%%%%%%%%%%%%%%%%%%%%%%%%%%%%%%%%%%%%%%%%%%%%%%%%%%%%%%%%%%%%%%%%
%%%%%%%%%%%%%%%%%%%%%%%%%%%%%%%%%%%%%%%%%%%%%%%%%%%%%%%%%%%%%%%%%%%%%%%%%%%

After tracing a grid of rays that is uniformly sampled on the pupil, we obtain the wavefront distribution at the exit pupil plane. Figure \ref{fig:Coherent Superposition} shows the wavefront distribution $\sum$ of converging light at the virtual exit pupil plane. We note that each ray of this wavefront distribution represents a spherical wavelet with a particular amplitude and phase, so the complex amplitude of Huygens' wavelet in the pupil plane can be represented by:

\begin{equation}\label{eqt:complex amplitude on pupil plane}
    \mathbf{E}(l_{x'y'}) = a_{0}\dfrac{\mathrm{e}^{\mathrm{i}k\mathbf{l_{x'y'}}}}{\mathbf{l_{x'y'}}},
\end{equation}
where $\mathbf{l_{x'y'}}$ is the length of total optical path from $(x_{0}, y_{0}, z_{0})$ to $(x', y', z')$, where $x'$ and $y'$ are the spatial coordinates of the pupil plane. $a_{0}$ is the amplitude of the spherical wave at the unit distance from the source. $k=2\pi/\lambda$ is the wave number of the light.

The complex amplitude propagating to the image plane can be formulated as follows:
\begin{equation}\label{eqt:complex amplitude on image plane}
    \mathbf{E}(\mathbf{l_{x'y'}}, \mathbf{r_{x''y''}}, \mathbf{K}) = a_{0}\dfrac{\mathrm{e}^{\mathrm{i}k\mathbf{l_{x'y'}}}}{\mathbf{l_{x'y'}}}\dfrac{\mathrm{e}^{\mathrm{i}k\mathbf{r_{x''y''}}}}{\mathbf{r_{x''y''}}}\mathbf{K}(\mathbf{D_{x'y'}}, \mathbf{r_{x''y''}}, n),
\end{equation}
where the $\mathbf{r_{x''y''}} = (x''-x', y''-y', z''-z')$, and $\mathbf{K}(\mathbf{D_{x'y'}}, \mathbf{r_{x''y''}}, n)$ is the obliquity factor of the spherical wavelet, which is defined as follows:

\begin{equation}\label{eqt:tilt factor}
    \mathbf{K}(\mathbf{D_{x'y'}}, \mathbf{r_{x''y''}}, \mathbf{n}) = \frac{1}{2}[cos<\mathbf{n}, \mathbf{r_{x''y''}}> - cos<\mathbf{n}, \mathbf{D_{x'y'}}>],
\end{equation}
where $\mathbf{n}$ is the normal unit vector of the exit pupil plane and $cos<\cdot, \cdot>$ is the operation of computing the cosine value of the two vectors. The relationships of $\mathbf{n}$, $\mathbf{D_{x'y'}}$, and $\mathbf{r_{x''y''}}$ are magnified in Figure \ref{fig:Coherent Superposition}.

Therefore, the complex amplitude $\mathbf{\widetilde{E}}_{x''y''}$ at the point $(x'', y'')$ on the image surface is the complex sum of all the wavefront distributions propagated from the exit pupil plane, which can be formulated as the following equation:
\begin{equation}\label{eqt:sum of complex amplitude}
    \mathbf{\widetilde{E}}_{x''y''} = \sum_{y'}\sum_{x'}\mathbf{E}(\mathbf{l_{x'y'}}, \mathbf{r_{x''y''}}, \mathbf{K}(\mathbf{D_{x'y'}}, \mathbf{r_{x''y''}}, \mathbf{n})),
\end{equation}
The diffraction intensity $I_{x''y''}$ can be expressed by:
\begin{equation}\label{eqt:superposition intensity}
    I_{x''y''} = \mathbf{\widetilde{E}}_{x''y''} .* \mathbf{\widetilde{E}^{*}}_{x''y''},
\end{equation}
Here the corresponding elements of the complex amplitude matrix $\mathbf{\widetilde{E}}_{x''y''}$ and $\mathbf{\widetilde{E}^{*}}_{x''y''}$ are multiplied. $\mathbf{\widetilde{E}^{*}}_{x''y''}$ is the complex conjugate of $\mathbf{\widetilde{E}}_{x''y''}$, and $I_{x''y''}$ is the resulting PSF matrix. Algorithm \ref{alg:PSF calculation} outlines all the steps of the optical PSF calculation. To keep the paper reasonably concise, a brief description of the raytracing is provided in Algorithm \ref{alg:PSF calculation}. We refer the readers to the work by \cite{smith2008modern} for an in-depth discussion about raytracing. Moreover, computing the optical PSF of the whole image plane is meaningless because the complex amplitudes of many sampled points are almost zero. Therefore, as shown in Algorithm \ref{alg:PSF calculation}, we first trace the main ray to acquire the imaging center $(x''_{c}, y''_{c}, z''_{c})$ on the image plane and then perform the optical PSF calculation within the sampling range $x''_{r}\times y''_{r}$.

Altogether, we illustrate each step of the optical PSF model in Section. \ref{sec:Ray tracing} and Section. \ref{sec:coherent superposition}. Different from traditional raytracing or FFT \cite{heckbert1995fourier} to calculate PSF, we regard each light arriving at the pupil plane as a secondary wavelet. Because they are emitted by a coherent object point $\mathcal{S}$, the wavelets will interfere with each other when the wavefront propagates to the image plane. Therefore, the PSF of a coherent object point $\mathcal{S}$ is the diffraction result of the pupil plane, where the complex amplitude of each sampled wavelet must be considered. In the lower-right corner of Figure \ref{fig:Coherent Superposition}, we show the different FOVs' energy PSFs computed by our method. However, because of the ISP systems in digital cameras, the computed optical PSF, which represents the energy diffusion on the imaging plane, cannot be directly used for imaging simulation. Therefore, in the following section, we illustrate how to integrate optical PSF into the imaging simulation system.

\subsection{Wavelength Response and Relative Illumination}
\label{sec:Wavelength Response and Relative Illumination}

Each pixel in a conventional camera sensor is covered by a single red, green, or blue color filter \cite{brooks2019unprocessing}. These color filters have different responses to light waves of different wavelengths, which can be formulated as $\mathit{C_{wav}(\lambda)}$. Moreover, in the imaging formation process, the off-axis field will suffer a loss of illumination, which is caused by lens shading and can be formulated as $\mathit{C_{ill}(fov, \lambda)}$. Therefore, for a precise simulation, the PSFs of different wavelengths and FOVs are multiplied by the specific distribution of wavelength responses and lens shading parameters. The influence of the wave response and relative illumination can be easily modeled by a coefficient:
\begin{equation}\label{wave response and relative illumination}
    \mathcal{C}_{e}(fov,\lambda) = \mathcal{C}_{ill}(fov,\lambda)\cdot\mathcal{C}_{wav}(\lambda),
\end{equation}
Here $fov$ is the normalized FOV, and $\lambda$ is the wavelength of object point $\mathcal{S}$. 
%%%%%%%%%%%%%%%%%%%%%%%%%%%%%%%%%%%%%%%%%%%%%%%%%%%%%%%%%%%%%%%%%%%%%%%%%%%%%
\subsection{Imaging Simulation Pipeline}

%%%%%%%%%%%%%%%%%%%%%%%%%%%%%%%%%%%%%%%%%%%%%%%%%%%%%%%%%%%%%%%%%%%%%%%%%%%%%
%%%%%%%%%%%%%%%%%%%%%%%%%%%%%%%%%%%%%%%%%%%%%%%%%%%%%%%%%%%%%%%%%%%%%%%%%%%%%
\begin{figure*}[ht]
  \centering
  \includegraphics[width=\linewidth]{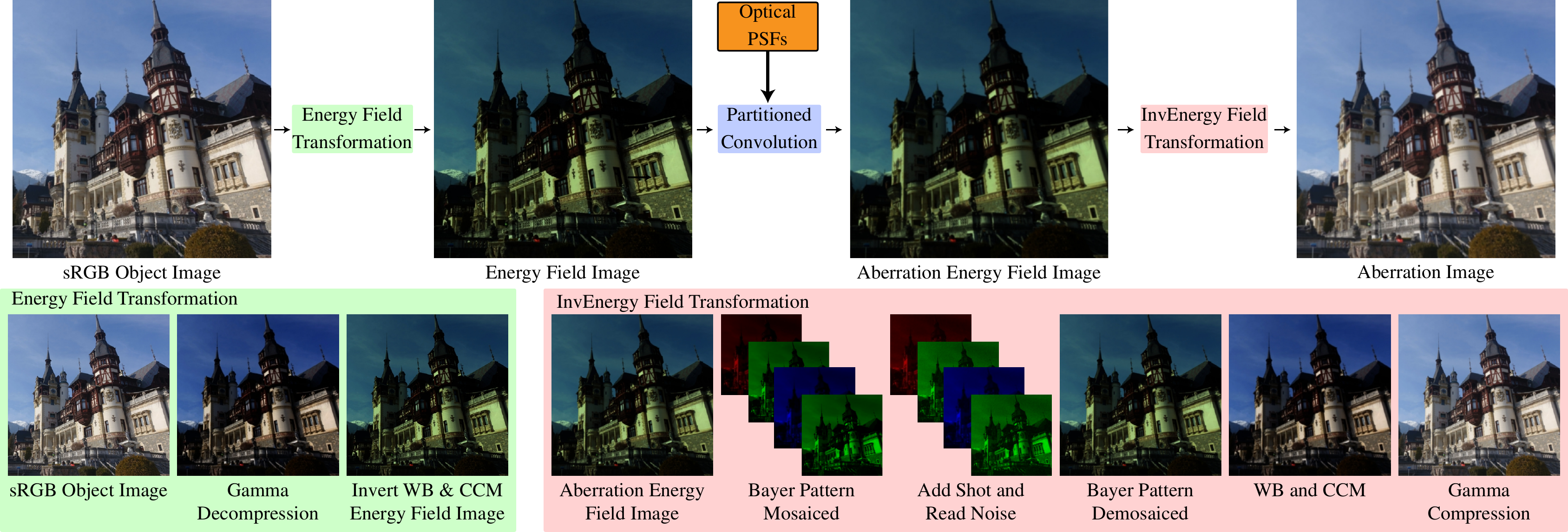}
  \caption{\textbf{Simulation pipeline of the imaging simulation system.} The input image is converted to the energy domain first. Then, the optical PSFs reunited in Section \ref{sec:Wavelength Response and Relative Illumination} are applied to the energy domain image in a partitioning convolution manner. Finally, the degraded image is transformed into raw Bayer data and converted to aberration images. The details of the energy domain transformation and inverted energy domain transformation are shown at the bottom.}
  \Description{Imaging Simulation System.}
  \label{fig:Imaging Simulation System}
\end{figure*}
%%%%%%%%%%%%%%%%%%%%%%%%%%%%%%%%%%%%%%%%%%%%%%%%%%%%%%%%%%%%%%%%%%%%%%%%%%%%%
%%%%%%%%%%%%%%%%%%%%%%%%%%%%%%%%%%%%%%%%%%%%%%%%%%%%%%%%%%%%%%%%%%%%%%%%%%%%%
Owing to the differences between the dynamic range of raw sensor data and the sensitivity of the humans eye, imaging signal processing systems are typically used to convert the raw sensor energy data to photographs \cite{hasinoff2016burst}. In our imaging simulation pipeline, we transform the natural object image $L$ to the energy domain data $L_{e}$ with the energy domain transformation, which is shown in Figure \ref{fig:Imaging Simulation System}. Specifically, for a general natural image, we first perform gamma decompression with the standard gamma curve, while fixing the input to the gamma curve with $\epsilon = 10^{-8}$ to prevent numerical instability. We refer the readers to the work by \cite{brooks2019unprocessing} for a more in-depth discussion of gamma decompression. After gamma decomposition, we convert the sRGB image to camera RGB data with the inverse of the fixed $3 \times 3$ color correction matrix (CCM). The CCM is determined by the camera sensor. Then, the white balance (WB) of the image is inverted by dividing the white balance value at random color temperature. The WB values of different color temperatures are obtained by calibration. In this way, we obtain the energy domain image $L_{e}$.

Based on the optical PSFs $I_{x''y''}$ calculated in Section. \ref{sec:Optical PSF Model}, we degrade the energy domain image $L_{e}$ with partitioned convolution. Partitioned convolution can be divided into three parts: Firstly we crop the image into uniform small pieces, then these pieces are convoluted with the PSFs of the corresponding FOVs, and finally the degraded pieces are spliced together. It is worth mentioning that the optical PSF calculated in Algorithm \ref{alg:PSF calculation} is the PSF of one point $(x_{0}, y_{0}, z_{0})$. To precisely simulate the imaging results, we uniformly sample the object plane and perform the PSF calculation in Algorithm \ref{alg:PSF calculation} on every sampled point. The overall PSF can be represented by $I(h, w, \lambda)$. Here, $h$, $w$ are the coordinates on the sensor plane (corresponding to the sampled point on the object plane), and $\lambda$ is the wavelength of the light source. For a captured image with $n\times m$ resolution, let $J_{e}$, $L_{e}$, $N_{e}$ $\in\mathbb{R}^{n\times m}$ be the observed energy image, the underlying sharp image, and the measurement noise image, respectively. The formation of the blurred observation $J_{e}$ can then be formulated as:

\begin{equation}\label{image formation model}
J_{e}(h, w) = \int \mathcal{C}_{e}(fov,\lambda)\cdot [I(h,w,\lambda)*L_{e}(h,w)]d\lambda + N_{e}(h, w),
\end{equation}
Here, $fov$ can be calculated by $(h, w)$ and the full FOV of the lens. It is important to note that $J_{e},L_{e}$ are raw-like images that indicate the energy received by the sensor.

After generating the blurred observation $J_{e}$ with Eq.(\ref{image formation model}), we invert the energy domain transformation by sequentially applying the operations in the lower-right corner of Figure \ref{fig:Imaging Simulation System}. To be more specific, because the red, green, and blue color filters are arranged in a Bayer pattern, such as R-G-G-B, we omit two of its three color values according to the Bayer filter pattern of the camera. After acquiring the mosaiced raw energy data, we approximate the shot and read noise together as a single heteroscedastic Gaussian distribution and add it to each channel of the raw image \cite{brooks2019unprocessing}. We refer the readers to \cite{brooks2019unprocessing} for details on synthetic noise implementation. Moreover, we sequentially apply the variation in the AHD demosaic algorithm \cite{lian2007adaptive}, white balance, color correction matrix and the inverse of the gamma operator to $J_{e}$. The pipeline of the imaging simulation system is shown in Figure \ref{fig:Imaging Simulation System}.

In conclusion, there are two significant differences between our proposed method and the imaging simulation methods that have been used in $Zemax^{\circledR}$ or $CODEV^{\circledR}$. First, our method calculates the superposition of every spherical wavelet, where the phase term of the ray is considered. Therefore, the proposed method can generate more accurate PSFs when compared with the traditional raytracing or FFT methods. Second, we combine the sensor information and the ISP parameters into the imaging simulation system. In this way, our method not only simulates the imaging process from the physical process but also obtains very similar imaging results visually.
%%%%%%%%%%%%%%%%%%%%%%%%%%%%%%%%%%%%%%%%%%%%%%%%%%%%%%%%%%%%%%%%%%%%%%%%%%%%%
%%%%%%%%%%%%%%%%%%%%%%%%%%%%%%%%%%%%%%%%%%%%%%%%%%%%%%%%%%%%%%%%%%%%%%%%%%%%%
%%%%%%%%%%%%%%%%%%%%%%%%%%%%%%%%%%%%%%%%%%%%%%%%%%%%%%%%%%%%%%%%%%%%%%%%%%%%%
\section{aberrations Image Recovery}
\label{sec:aberrations Image Recovery}
Due to the spatially variant properties of optical aberrations, a spatial-adaptive recovery framework is constructed to restore the degraded image. In this section, we first explain the network architecture and the additional input. Then we illustrate the deformable ResBlock which changes the shapes of convolutional kernels according to the related FOV information. Third, we introduce the context block inserted in the minimum scale of our model. The input used to train our network is the synthetic data that has been processed using the proposed imaging simulation system. In summary, this neural network model combines the optical parameters of specific lens designs, and has spatial adaptiveness to handle aberrations varying with the FOV.
%%%%%%%%%%%%%%%%%%%%%%%%%%%%%%%%%%%%%%%%%%%%%%%%%%%%%%%%%%%%%%%%%%%%%%%%%%%%%

\subsection{Recovery Framework}
%%%%%%%%%%%%%%%%%%%%%%%%%%%%%%%%%%%%%%%%%%%%%%%%%%%%%%%%%%%%%%%%%%%%%%%%%%%%%
%%%%%%%%%%%%%%%%%%%%%%%%%%%%%%%%%%%%%%%%%%%%%%%%%%%%%%%%%%%%%%%%%%%%%%%%%%%%%
\begin{figure*}[ht]
  \centering
  \includegraphics[width=\linewidth]{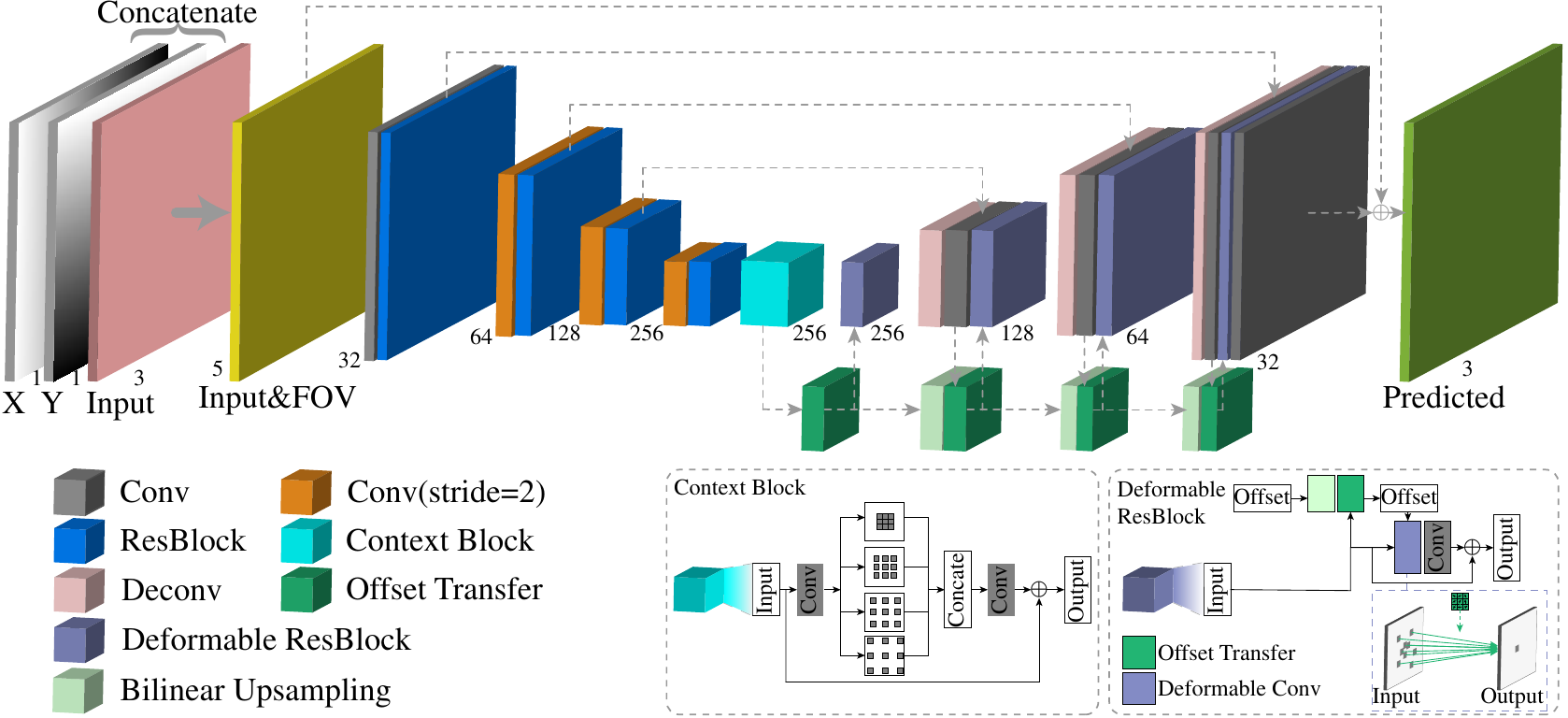}
  \caption{\textbf{The proposed neural network architecture.} The UNet-based network architecture is shown on the top and the layer configurations of encoder/decoder stages are illustrated with different colored blocks (bottom left). The deformable ResBlock and the context block are detailed in the bottom-right corner. The number of channels for each scale is marked with each block. In particular, we choose MSELoss as the loss function when training, which evaluates the fidelity between the predicted image and the ground truth.}
  \Description{Network Architecture.}
  \label{fig:Network Architecture}
\end{figure*}
%%%%%%%%%%%%%%%%%%%%%%%%%%%%%%%%%%%%%%%%%%%%%%%%%%%%%%%%%%%%%%%%%%%%%%%%%%%%%
%%%%%%%%%%%%%%%%%%%%%%%%%%%%%%%%%%%%%%%%%%%%%%%%%%%%%%%%%%%%%%%%%%%%%%%%%%%%%
We propose a novel network for the retrieval of the latent image $L$ from corrupted data $J$. Because we generate the training data by simulation, the input and the ground truth of our network are pixel-to-pixel aligned. This characteristic greatly improves the stability of training. In addition, the aberrations have a high correlation with the FOV, and we compute the $x$ and $y$ coordinates of each pixel and concatenate the FOV image with the aberration inputs (as shown in Figure \ref{fig:Network Architecture}). This operation enables our network to directly perceive the FOV information, which can be regarded as another kind of natural prior. The proposed network is shown in Figure \ref{fig:Network Architecture}. Specifically, we adopt a variant of the UNet architecture \cite{ronneberger2015u}. 

In regard to the loss function, the strictly paired data for supervised training allow us to use fidelity loss to learn robust texture details, thereby preventing fake information in the recovered image $L$. Specifically, the fidelity loss function is:

\begin{equation}\label{fidelity loss function}
    L(\theta) = \frac{1}{N} \sum \limits^{N}_{n=1} \Vert Net(I_{n}^{ab}) - I_{n}^{gt} \Vert^{2}_{2}.
\end{equation}

where $\theta$ denotes the learned parameters in the network. $I_{n}^{ab}$ are the aberration inputs, and $I_{n}^{gt}$ are the corresponding ground truths of $I_{n}^{ab}$. We note that many loss functions (such as perceptual loss \cite{johnson2016perceptual} or GAN loss \cite{ledig2017photo}) will generate sharper visual results on natural images. These loss functions also introduce fake texture to the restorations and are not good at suppressing noise. Because the noise level of mobilephone varies greatly while illumination and color temperature are changing. Therefore, we adopt the MSELoss, which is robust to different noise levels.
%%%%%%%%%%%%%%%%%%%%%%%%%%%%%%%%%%%%%%%%%%%%%%%%%%%%%%%%%%%%%%%%%%%%%%%%%%%%%
%%%%%%%%%%%%%%%%%%%%%%%%%%%%%%%%%%%%%%%%%%%%%%%%%%%%%%%%%%%%%%%%%%%%%%%%%%%%%
\subsection{Deformable ResBlock}
When the field of view increases, the higher-order aberration coefficient deviates from that computed by the Gaussian paraxial model. The different sizes and shapes of off-axis PSFs are caused by these deviations. However, the traditional convolution operation strictly adopts the feature of fixed locations around its center, which will introduce irrelevant features into the output and discard the information truly related to PSFs.

The size of optical PSFs ranges from 100 pixels to 1,600 pixels (squared), which means that our network must extract the features based on the FOV information. Therefore, we introduce the deformable convolution in the decorder path \cite{dai2017deformable, zhu2019deformable}. While different from the method in \cite{chang2020spatial}, we directly concatenate the input and the upsampled offset. After the concatenation, we refine the output offset with a convolution layer. In this way, the shape and the size of the convolution kernel can be changed adaptively according to the field-of-view information. 

%%%%%%%%%%%%%%%%%%%%%%%%%%%%%%%%%%%%%%%%%%%%%%%%%%%%%%%%%%%%%%%%%%%%%%%%%%%%%
%%%%%%%%%%%%%%%%%%%%%%%%%%%%%%%%%%%%%%%%%%%%%%%%%%%%%%%%%%%%%%%%%%%%%%%%%%%%%
\subsection{Context block}
Multiscale information is important for handling large PSFs. However, multiple downsamplings will increase the overhead of the network and destroy image structure. Therefore, we introduce a context block \cite{chang2020spatial} into the minimum scale between encoder and decoder, which increases the receptive field and reconstructs multiscale information without further downsampling. As shown in Figure \ref{fig:Network Architecture}, several dilated convolutions are used to extract features from different receptive fields, which are concatenated later to estimate the output. 

%%%%%%%%%%%%%%%%%%%%%%%%%%%%%%%%%%%%%%%%%%%%%%%%%%%%%%%%%%%%%%%%%%%%%%%%%%%%%

\section{Data Preparation}
\label{sec:Data Preparation}
%%%%%%%%%%%%%%%%%%%%%%%%%%%%%%%%%%%%%%%%%%%%%%%%%%%%%%%%%%%%%%%%%%%%%%%%%%%%%
%%%%%%%%%%%%%%%%%%%%%%%%%%%%%%%%%%%%%%%%%%%%%%%%%%%%%%%%%%%%%%%%%%%%%%%%%%%%%
\begin{figure}[t]
  \centering
  \includegraphics[width=\linewidth]{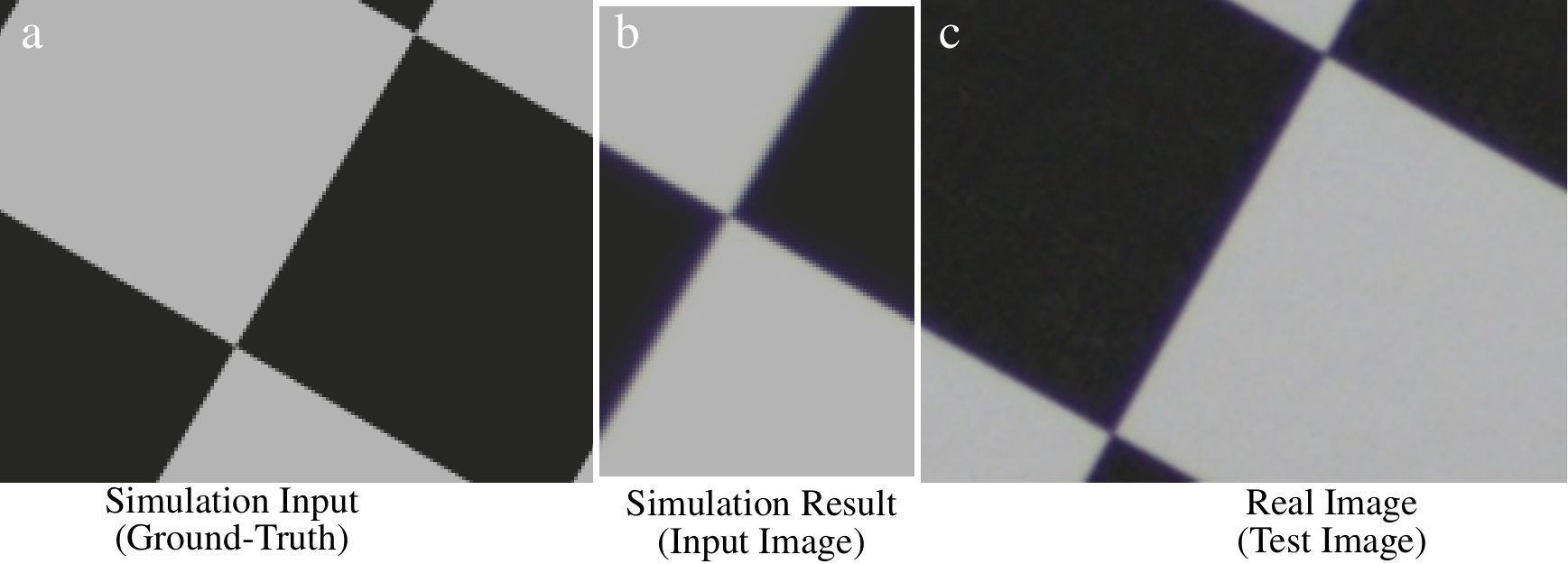}
  \caption{\textbf{Illustration of data preparation} In imaging simulation, (a) is the input, (b) is the synthetic result and (c) is the real image taken by the DSLR camera. In network training, (a) is the ground truth, (b) is the input data and (c) is the image in the test datasets.}
  \Description{data generation}
  \label{fig:data generation}
\end{figure}
%%%%%%%%%%%%%%%%%%%%%%%%%%%%%%%%%%%%%%%%%%%%%%%%%%%%%%%%%%%%%%%%%%%%%%%%%%%%%
%%%%%%%%%%%%%%%%%%%%%%%%%%%%%%%%%%%%%%%%%%%%%%%%%%%%%%%%%%%%%%%%%%%%%%%%%%%%%
How to acquire targeted data pairs that follow the distribution of real scenes is the principle part of the supervised learning method. Our proposed method does not need any shooting or registration operation, which is a practical method for correcting the optical aberrations of various commercial lenses. In this section, we illustrate how we generate the data pairs of our networks.

\subsection{Simulate Training Data}
\label{sec:Simulate Training Data}
As shown in Figure \ref{fig:data generation}, the simulation result of our simulation system and the corresponding digital image are the input and ground truth of our network, respectively. In the partitioned convolution operation in Figure \ref{fig:Imaging Simulation System}, we calculate 240,000 ($400 \times 600$) PSFs for the customized DSLR camera lens and 120,000 ($300 \times 400$) PSFs for HUAWEI HONOR 20. The number of calculated PSFs is decided by the resolution of the sensor, {\itshape i.e.}, the resolution of the sensor in the DSLR camera is $4,000 \times 6,000$, and we compute PSFs for every $10 \times 10$ patch. We adopt DIV2K \cite{martin2001database} which contains 800 images of 2K resolution as the digital images to generate training data pairs.

In regard to the PSF calculation, we set the fixed object distance $z_{0}$ for each camera. The shooting distance is $1750mm$ for the DSLR camera, and it is $600mm$ for HUAWEI HONOR 20. For the DSLR camera, we compute 31 PSFs with wavelengths varying from 400nm to 700nm with a 10nm interval. For HUAWEI HONOR 20, we calculate 34 PSFs varying from 400nm to 730nm. The wave distribution of PSFs is determined by the upper limit and the bottom limit of the sensor's wave response. Finally, the PSFs $I(h, w, \lambda)$ are reunited by the wave response functions of the sensor and lens shading parameters $\mathcal{C}_{e}(fov, \lambda)$.

\subsection{Generate Testing Images}
\label{sec:Generate Testing Images}
To evaluate the effect of our trained network, we photograph 100 images with a DSLR camera and HUAWEI HONOR 20, respectively. After acquiring the raw data taken by the camera, we sequentially apply the steps in Figure \ref{fig:Overview}.b to obtain the sRGB image. These sRGB photographs make up the testing datasets. We note that the sharpen operation in classical ISP will affect optical aberration degradation, therefore, we use the self-defined postprocessing pipeline to transform raw data into sRGB images.

\subsection{Training Details}
\label{sec:Training Details}
In this way, we construct the data pairs for our network. In our network architecture, the channel numbers of each layer are marked at the bottom of the corresponding blocks (as shown in Figure \ref{fig:Network Architecture}). In terms of the hyperparameter of training, our settings are basically the same as \cite{chang2020spatial}. In addition, the $H, W$ of our training input is $256\times256$, which allows the network to receive a large amount of spatially variant degradation.

%%%%%%%%%%%%%%%%%%%%%%%%%%%%%%%%%%%%%%%%%%%%%%%%%%%%%%%%%%%%%%%%%%%%%%%%%%%%%

\section{Analysis}
\label{sec:Analysis}

In this section, we first analyze the PSF characteristics of the customized DSLR camera lens and HUAWEI HONOR 20 (\textbf{the extrinsic parameters of optical designs are listed in the supplementary file}). Second, comprehensive experiments are conducted to prove the accuracy of the imaging simulation system. Here, the experiments are divided into two parts: one part validates the accuracy of the calculated PSFs, and the other part compares the simulation results with the results generated by Zemax and the photograph taken by the camera. Third, we perform the ablation study of our network architecture. Finally, we compare the proposed method with the state-of-the-art deblurring methods, and the evaluation metrics are PSNR and SSIM. 

\subsection{Optical PSF Analysis}
\label{sec:Optical PSF Analysis}
%%%%%%%%%%%%%%%%%%%%%%%%%%%%%%%%%%%%%%%%%%%%%%%%%%%%%%%%%%%%%%%%%%%%%%%%%%%%%
%%%%%%%%%%%%%%%%%%%%%%%%%%%%%%%%%%%%%%%%%%%%%%%%%%%%%%%%%%%%%%%%%%%%%%%%%%%%%
\begin{figure}[t]
  \centering
  \includegraphics[width=\linewidth]{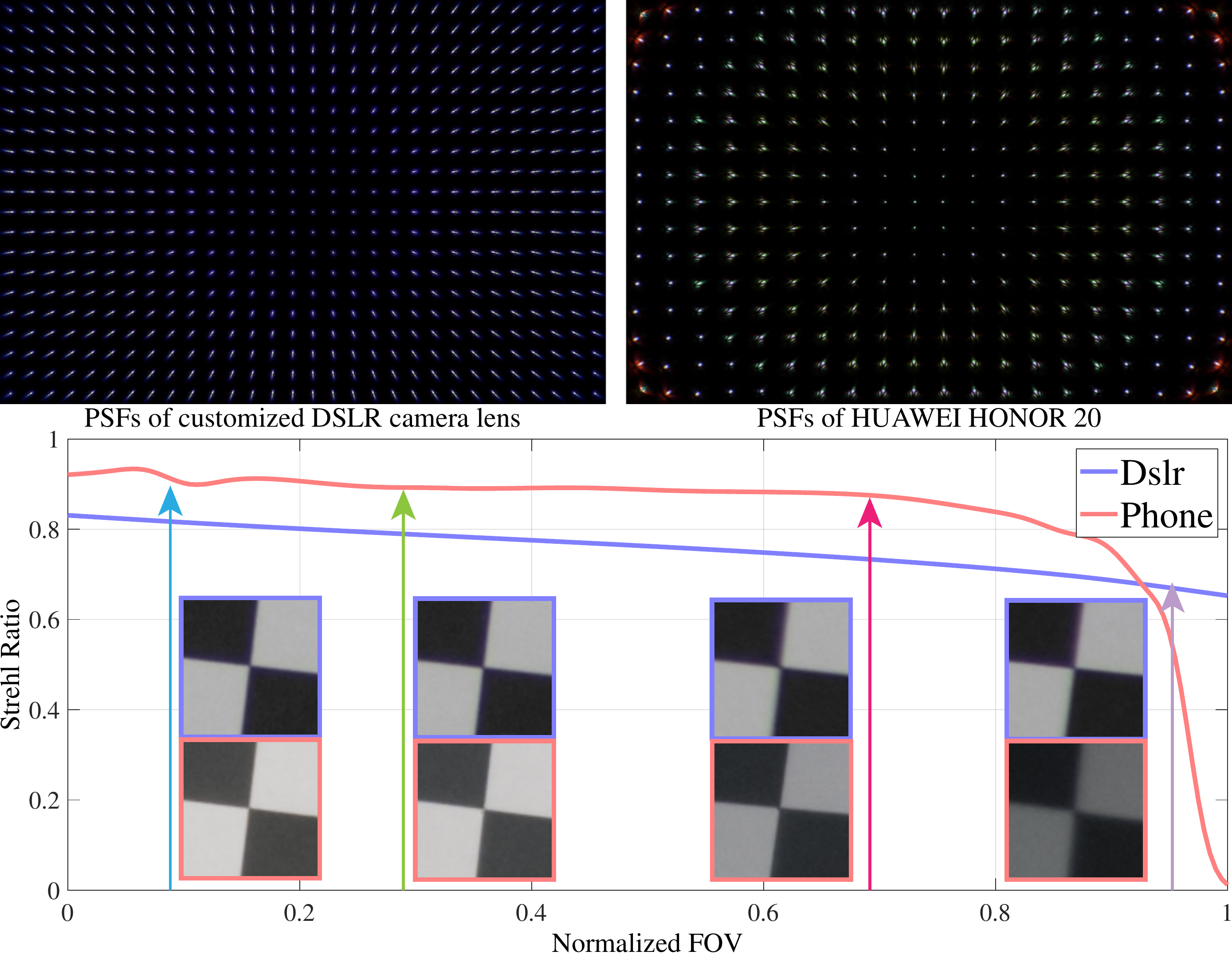}
  \caption{\textbf{PSFs characteristics.} The patchwise PSF distributions of the customized DSLR camera lens and HUAWEI HONOR 20 are sketched on top. The bottom panel shows the Strehl ratio \cite{mahajan1982strehl} curves of the DSLR camera and HUAWEI HONOR 20. The real measurements of different FOVs are plotted by side, where the blue boxes show the DSLR results and the orange boxes show the images taken by HUAWEI HONOR 20. These results demonstrate the spatially variant characteristics of PSFs.}
  \Description{PSF characteristic.}
  \label{fig:PSF characteristic}
\end{figure}
%%%%%%%%%%%%%%%%%%%%%%%%%%%%%%%%%%%%%%%%%%%%%%%%%%%%%%%%%%%%%%%%%%%%%%%%%%%%%
%%%%%%%%%%%%%%%%%%%%%%%%%%%%%%%%%%%%%%%%%%%%%%%%%%%%%%%%%%%%%%%%%%%%%%%%%%%%%

%%%%%%%%%%%%%%%%%%%%%%%%%%%%%%%%%%%%%%%%%%%%%%%%%%%%%%%%%%%%%%%%%%%%%%%%%%%%%
The optical PSF distributions of the customized DSLR camera lens and HUAWEI HONOR 20 are sketched on the top of Figure \ref{fig:PSF characteristic}, which are centrally symmetric and have a high correlation with the FOV. We note that the optical PSFs of HUAWEI HONOR 20 reveal a wide range of spatial variation over the region, while the PSFs of the customized DSLR camera are relatively consistent. Our network architecture takes advantage of these characteristics. It engages with the FOV information to perform spatial adaptive reconstruction for degraded images. It is worth mentioning that our method directly calculates the resulting PSFs, which is different from the optimization method of PSF estimation \cite{mosleh2015camera}.

We plot the Strehl ratio \cite{mahajan1982strehl} curves and the measured patches in $0.1$, $0.3$, $0.7$, and $0.96$ FOVs of two optical designs, see the bottom of Figure \ref{fig:PSF characteristic}. The Strehl ratio can be used to evaluate the energy diffusion in different FOVs, which has some correlation to the degradation of the image. As shown in Figure \ref{fig:PSF characteristic}, the Strehl ratio turns lower from the center to the whole field, resulting in more blurred photography. Moreover, unlike the customized DSLR camera lens, the shape and size of PSFs in HUAWEI HONOR 20 abruptly change in regard to the $0.9$ FOV, which is common to camera lenses. Thus, the PSF characteristics of different devices are totally different, which means that it is significant to generate targeted datasets for a specific lens.

\subsection{Authenticity of Imaging Simulation System}
\label{sec:Authenticity of Imaging Simulation System}
%%%%%%%%%%%%%%%%%%%%%%%%%%%%%%%%%%%%%%%%%%%%%%%%%%%%%%%%%%%%%%%%%%%%%%%%%%%%%
%%%%%%%%%%%%%%%%%%%%%%%%%%%%%%%%%%%%%%%%%%%%%%%%%%%%%%%%%%%%%%%%%%%%%%%%%%%%%
\begin{figure}[t]
  \centering
  \includegraphics[width=\linewidth]{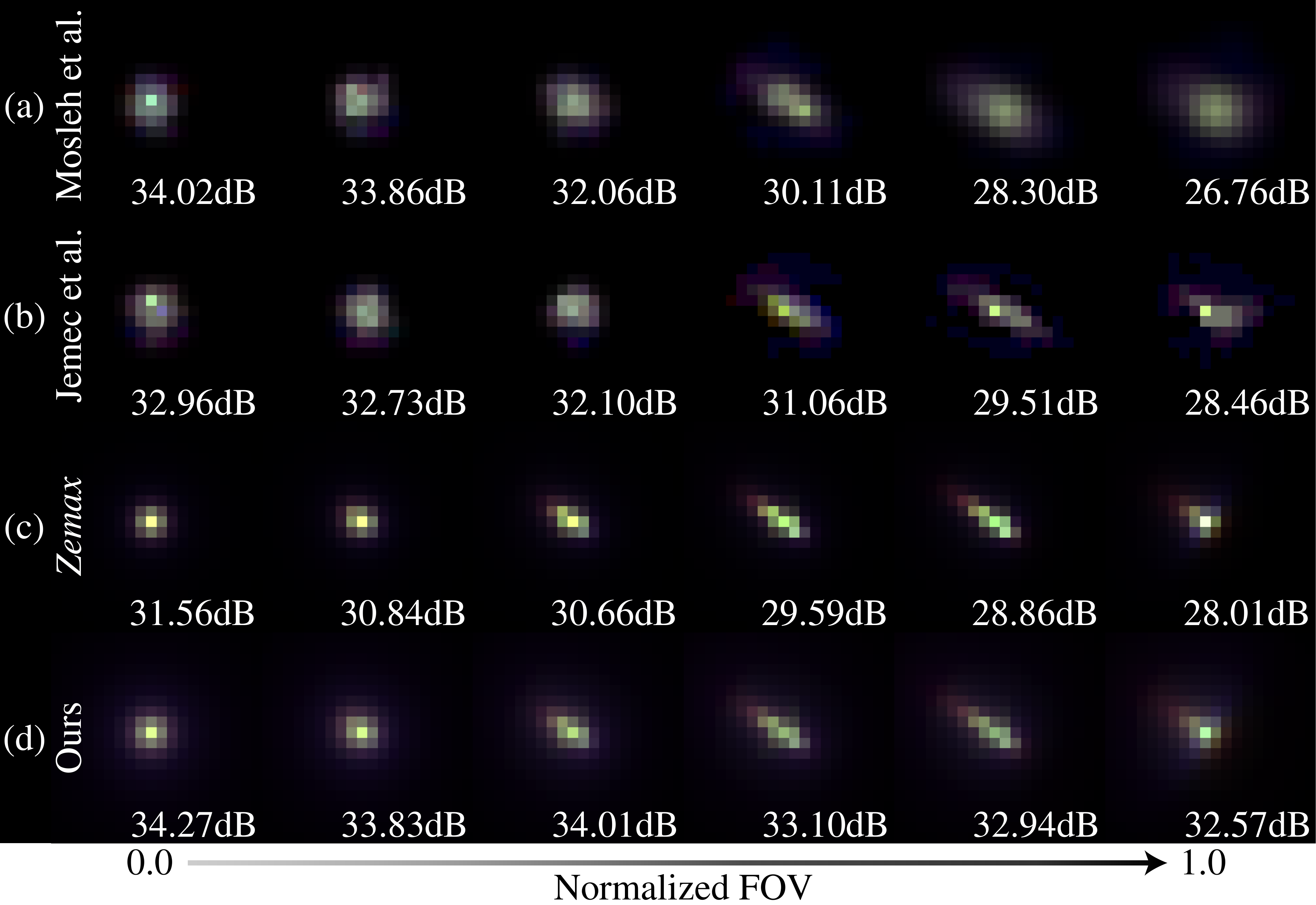}
  \caption{\textbf{Authenticity of PSFs.} Here, we show the PSFs calculated by different methods, which are arranged from left to right according to the corresponding FOV. (a) shows the PSFs optimized by the method in \cite{mosleh2015camera} (we use the chart of the noise pattern). (b) is the PSFs calibrated by the virtual point-like source \cite{jemec20172d}. (c) is the PSFs calculated by $Zemax^{\circledR}$'s raytracing. (d) shows the results of the proposed method. The $PSNRs$ of these PSFs' imaging simulation (compared to the real image) are noted in the bottom-right corner (we use $100\times 100$ checkerboard for $PSNR$ calculation).}
  \Description{Authenticity of PSFs.}
  \label{fig:Authenticity of PSFs}
\end{figure}
%%%%%%%%%%%%%%%%%%%%%%%%%%%%%%%%%%%%%%%%%%%%%%%%%%%%%%%%%%%%%%%%%%%%%%%%%%%%%
%%%%%%%%%%%%%%%%%%%%%%%%%%%%%%%%%%%%%%%%%%%%%%%%%%%%%%%%%%%%%%%%%%%%%%%%%%%%%

%%%%%%%%%%%%%%%%%%%%%%%%%%%%%%%%%%%%%%%%%%%%%%%%%%%%%%%%%%%%%%%%%%%%%%%%%%%%%
%%%%%%%%%%%%%%%%%%%%%%%%%%%%%%%%%%%%%%%%%%%%%%%%%%%%%%%%%%%%%%%%%%%%%%%%%%%%%
\begin{figure}[t]
  \centering
  \includegraphics[width=\linewidth]{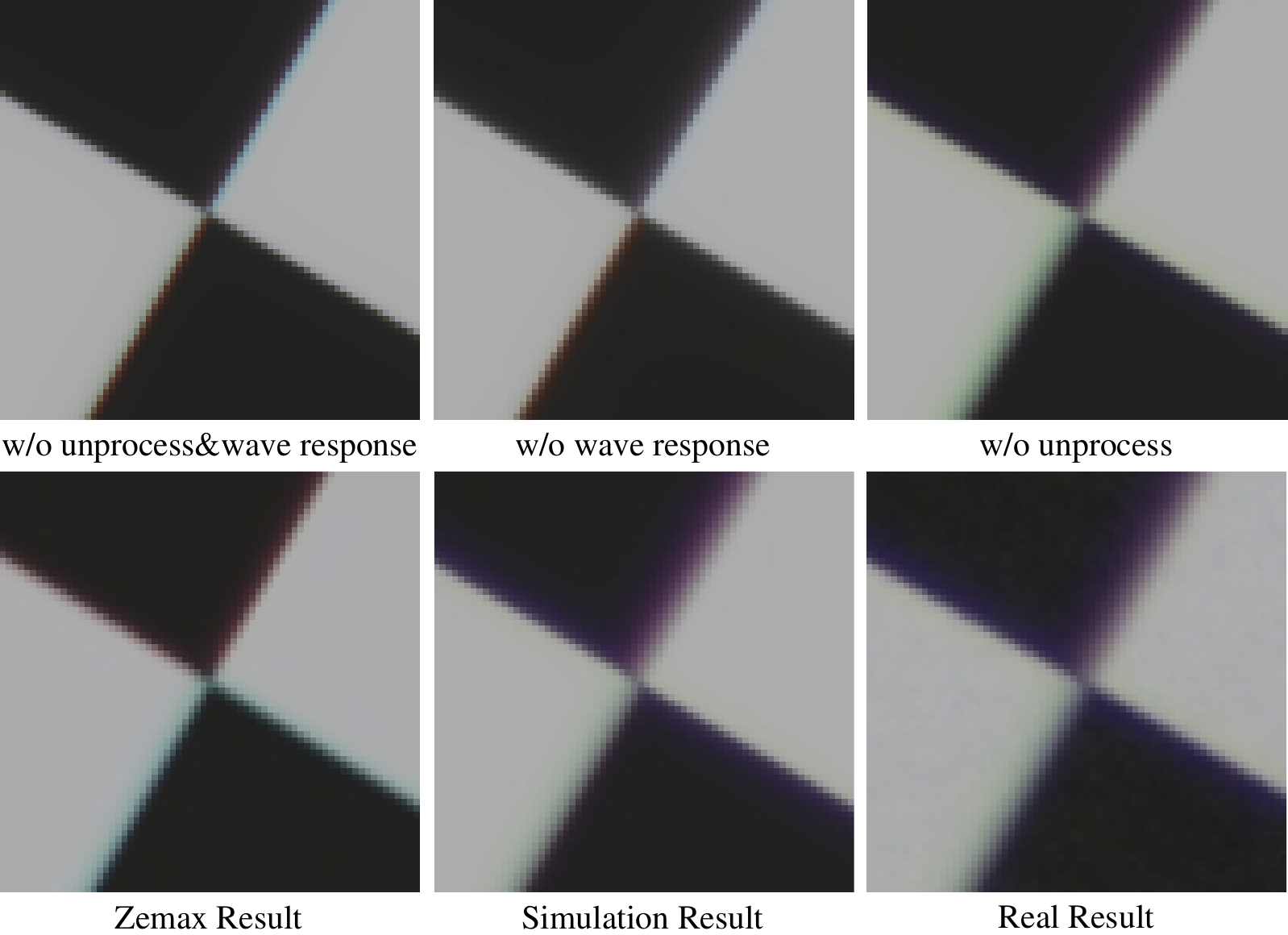}
  \caption{\textbf{Accuracy of Imaging Simulation.} In the first row, we show the ablation study of the imaging simulation system. In the second row we compare our result with the image generated by $Zemax^{\circledR}$ and the data taken by the DSLR camera.}
  \Description{Accuracy of Imaging Simulation.}
  \label{fig:Accuracy of Imaging Simulation}
\end{figure}
%%%%%%%%%%%%%%%%%%%%%%%%%%%%%%%%%%%%%%%%%%%%%%%%%%%%%%%%%%%%%%%%%%%%%%%%%%%%%
%%%%%%%%%%%%%%%%%%%%%%%%%%%%%%%%%%%%%%%%%%%%%%%%%%%%%%%%%%%%%%%%%%%%%%%%%%%%%

To prove the accuracy of the proposed imaging simulation system, we designed a two-pronged experiment. One prong is to verify the accuracy of the calculated PSFs: First, we calibrate the PSFs of the DSLR camera with the virtual point-like source \cite{jemec20172d}. Then, the PSFs of the same camera are estimated by the optimization method used in \cite{mosleh2015camera} (we use the chart of the noise pattern). Finally, we calculate the PSFs by $Zemax^{\circledR}$'s raytracing. The comparison is shown in Figure \ref{fig:Authenticity of PSFs}, and we postprocess all the PSFs by the same postprocessing operations for visualization. When the FOV increases, the PSFs estimated by \cite{mosleh2015camera} (Figure \ref{fig:Authenticity of PSFs}.(a)) are becoming more blurry and the $PSNRs$ decrease greatly. We note that from the center to the edge of the image, the relative illumination decreases a lot. Therefore, the edge's results of \cite{mosleh2015camera} are not ideal because the method has difficulty converging when the degraded information is masked by noise. The marginal PSFs calibrated by \cite{jemec20172d} (Figure \ref{fig:Authenticity of PSFs}.(b)) also suffer from this problem. Different form these methods that need to shoot targets, our method directly calculate the PSFs from the lens prescriptions and will not be affected by noise. By comparing the $PSNRs$ of the proposed method and $Zemax^{\circledR}$, we prove that our results are more accurate and more similiar to the real image. Moreover, these results also validate the necessity of the coherent superposition. 
%%%%%%%%%%%%%%%%%%%%%%%%%%%%%%%%%%%%%%%%%%%%%%%%%%%%%%%%%%%%%%%%%%%%%%%%%%%%%
%%%%%%%%%%%%%%%%%%%%%%%%%%%%%%%%%%%%%%%%%%%%%%%%%%%%%%%%%%%%%%%%%%%%%%%%%%%%%
\begin{table}
  \caption{\textbf{Quantitive comparison of MTF area} We assess the MTF area, which is the area between the MTF curve and the cycle per pixel axis, for comparison. The ratio is the result of the simulation MTF area divided by the real MTF area.}
  \label{tab:MTF comparison}
  \begin{tabular}{c|c|c|c}
    \hline
    FOV & Siml MTF Area & Real MTF Area & Ratio\\
    \hline
    0.0 & 0.25421 & 0.24805 & 1.02898\\
    0.3 & 0.24284 & 0.24285 & 0.99995\\
    0.5 & 0.22133 & 0.22122 & 1.00052\\
    0.7 & 0.22034 & 0.21804 & 1.01055\\
    0.9 & 0.21156 & 0.20553 & 1.02934\\
    \hline
\end{tabular}
\end{table}
%%%%%%%%%%%%%%%%%%%%%%%%%%%%%%%%%%%%%%%%%%%%%%%%%%%%%%%%%%%%%%%%%%%%%%%%%%%%%
%%%%%%%%%%%%%%%%%%%%%%%%%%%%%%%%%%%%%%%%%%%%%%%%%%%%%%%%%%%%%%%%%%%%%%%%%%%%%

Another method for verifying the authenticity of our method is to compare the simulation results. We shoot the checkerboard chart with the corresponding camera and simulate the imaging results with $Zemax^{\circledR}$, where the parameters used are the same as those in our simulation. Moreover, we also perform an ablation study on the proposed imaging simulation system. The visual comparison of the results is shown in Figure \ref{fig:Accuracy of Imaging Simulation}. Obviously, the imaging results of our method is the closest to the real data. We note that the image generated by $Zemax^{\circledR}$ has no combination of the sensor's spectral information and the ISP parameters, accurately. This might be the reason that $\textit{Zemax}^{\circledR}$ produces inaccurate results. In addition, we compare the modulation transfer function (MTF) \cite{schroeder1981modulation} in different FOVs, which shows the accuracy of the proposed method through the whole region of the sensor (as shown in Table \ref{tab:MTF comparison}). 

In summary, we prove the authenticity of the proposed imaging simulation system mainly from two aspects. First, we demonstrate that our PSF calculation method is more accurate than the optimization method. When compared with the calibration methods, the proposed method is more practical and will not be affected by noise. The comparison between $Zemax^{\circledR}$ and the proposed method proves the accuracy of our method and validates the necessity of coherent superposition. Second, we compare the simulation results with the image of $Zemax^{\circledR}$'s simulation and the photograph of the camera, which verifies the authenticity of the imaging simulation system and every component is necessary.

\subsection{Ablation Study}
\label{sec:Ablation Study}
%%%%%%%%%%%%%%%%%%%%%%%%%%%%%%%%%%%%%%%%%%%%%%%%%%%%%%%%%%%%%%%%%%%%%%%%%%%%%
%%%%%%%%%%%%%%%%%%%%%%%%%%%%%%%%%%%%%%%%%%%%%%%%%%%%%%%%%%%%%%%%%%%%%%%%%%%%%
\begin{table*}
  \caption{\textbf{Ablation study.} The evaluation metric PSNRs are based on the simulation datasets for customized DSLR camera lens.}
  \label{tab:Ablation Study}
  \begin{tabular}{c|c|c|c|c|c|c|c|c}
    \hline
    Field-Of-View Input &  & \checkmark &            &            & \checkmark & \checkmark &            & \checkmark\\
    Deformable ResBlock &  &            & \checkmark &            & \checkmark &            & \checkmark & \checkmark\\
    Context Block       &  &            &            & \checkmark &            & \checkmark & \checkmark & \checkmark\\
    \hline
    PSNR & 40.14 & 43.72 & 40.63 & 40.38 & 44.36 & 44.01 & 41.23 & 44.77\\
    \hline
\end{tabular}
\end{table*}
%%%%%%%%%%%%%%%%%%%%%%%%%%%%%%%%%%%%%%%%%%%%%%%%%%%%%%%%%%%%%%%%%%%%%%%%%%%%%
%%%%%%%%%%%%%%%%%%%%%%%%%%%%%%%%%%%%%%%%%%%%%%%%%%%%%%%%%%%%%%%%%%%%%%%%%%%%%

We perform the ablation study of our network architecture on the simulation dataset of the DSLR camera lens. The results are shown in Table \ref{tab:Ablation Study}, and the evaluation metric is PSNR. 

\textbf{Ablation on FOV Input} We concatenate the FOV image with the aberrations input image. It is obvious that the FOV image helps the network determine the spatially variant blurring and displacement of optical aberrations.

\textbf{Ablation on Deformable ResBlock} Without the deformable ResBlock, the network loses the spatial adaptability to handle different PSFs of optical aberrations. If we replace the deformable ResBlock with ResBlock, the performance of our networks significantly decreases, which indicates its effect.

\textbf{Ablation on Context Block} The context block helps the network capture larger field information while not damaging the structure of the feature map. The comparison validates its effectiveness.

\subsection{Recovery Comparison}
\label{sec:Recovery Comparison}
Because of the spatial variation over aberrant PSFs, we design a brand new architecture to perform spatially adaptive recovery on the degraded image. In this section, we compare our algorithm with state-of-the-art deblurring methods.
%%%%%%%%%%%%%%%%%%%%%%%%%%%%%%%%%%%%%%%%%%%%%%%%%%%%%%%%%%%%%%%%%%%%%%%%%%%%%
%%%%%%%%%%%%%%%%%%%%%%%%%%%%%%%%%%%%%%%%%%%%%%%%%%%%%%%%%%%%%%%%%%%%%%%%%%%%%
\begin{figure}[t]
  \centering
  \includegraphics[width=\linewidth]{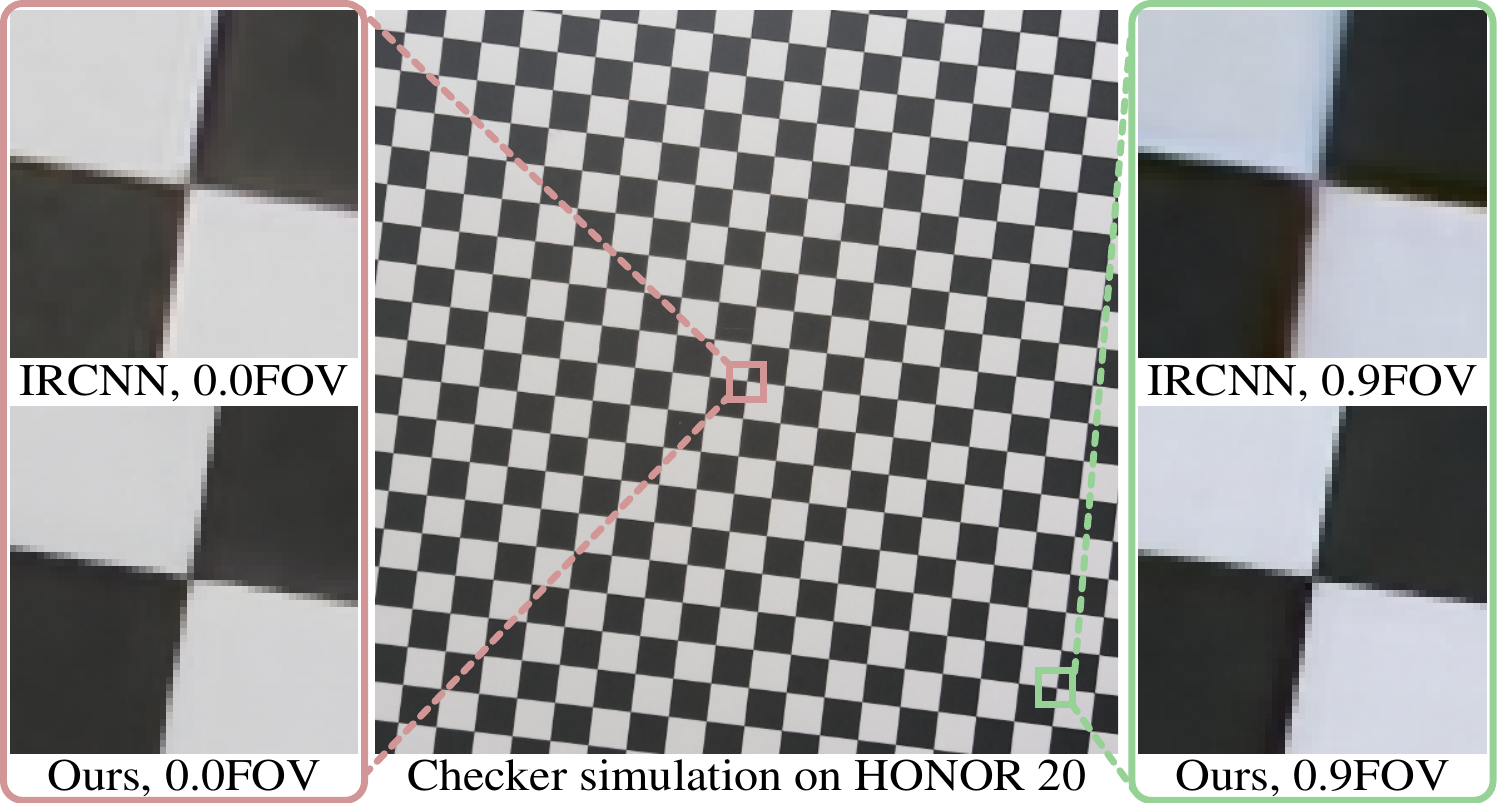}
  \caption{\textbf{Spatial adaptability of the proposed network.} We compare the different FOV restorations of IRCNN and ours. Obviously, IRCNN is not effective in dealing with the blur varying with the space.}
  \Description{Spatial Adaptive Illustration.}
  \label{fig:Spatial Adaptive Illustration}
\end{figure}
%%%%%%%%%%%%%%%%%%%%%%%%%%%%%%%%%%%%%%%%%%%%%%%%%%%%%%%%%%%%%%%%%%%%%%%%%%%%%
%%%%%%%%%%%%%%%%%%%%%%%%%%%%%%%%%%%%%%%%%%%%%%%%%%%%%%%%%%%%%%%%%%%%%%%%%%%%%

\subsubsection{Synthetic aberrations Image}
\label{sec:Synthetic aberrations Image}

To evaluate the effectiveness of the proposed approach, we select the IRCNN \cite{zhang2017learning}, GLRA \cite{ren2018deep} and Self-Deblur \cite{ren2020neural} as the representatives of the combination of model-based optimization methods and discriminative learning methods. Moreover, we choose some deep learning-based methods, such as SRN \cite{tao2018scale}, for comparisons. For a fair comparison, all the state-of-the-art methods employ the default setting provided by the corresponding authors. It is worth emphasizing that all these comparison methods are retrained by the synthetic datasets of the DSLR camera (DSLR) and HUAWEI HONOR 20 (Phone).

To prove the spatial adaptability of the proposed approach, we compare the restoration results with other deblur methods when evaluating synthetic data. As shown in Figure \ref{fig:Spatial Adaptive Illustration}, even though the Phone PSFs show a wide range of spatial variation over regions (as shown in Figure \ref{fig:PSF characteristic}), the proposed method performs spatial-adaptive recovery on the degraded input of different FOVs. IRCNN, which is designed for globally consistent blur, generates imperfect recovery results at the border of the checker. Because the edge of the FOV is much more blurred than the center of the image, the globally consistent deblur approach will receive a compromise between the center and the edge.
%%%%%%%%%%%%%%%%%%%%%%%%%%%%%%%%%%%%%%%%%%%%%%%%%%%%%%%%%%%%%%%%%%%%%%%%%%%%%
%%%%%%%%%%%%%%%%%%%%%%%%%%%%%%%%%%%%%%%%%%%%%%%%%%%%%%%%%%%%%%%%%%%%%%%%%%%%%
\begin{table}[t]
  \caption{\textbf{Quantitive comparison of synthetic aberrations correction.} The compared methods are SRN, GLRA, IRCNN, Self-Deblur, and our proposed network model. We assess PSNR, SSIM to complete the comparisons.}
  \label{tab:synthetic contrast}
  \begin{tabular}{c|c|ccccc}
    \hline
    Dataset                & Metric & SRN & GLRA & IRCNN & \makecell[c]{Self-\\Deblur} & Ours\\
    \hline\hline
    \multirow{2}{*}{DSLR}  & PSNR & 43.91 & 44.13 & 43.67 & 44.37 & \textbf{44.77}\\
    ~                      & SSIM & 0.996 & 0.994 & 0.992 & 0.996 & \textbf{0.998}\\
    \hline
    \multirow{2}{*}{Phone} & PSNR & 36.28 & 35.94 & 36.05 & 36.22 & \textbf{37.85}\\
    ~                      & SSIM & 0.980 & 0.978 & 0.975 & 0.978 & \textbf{0.986}\\
    \hline
    DSLR                   & PSNR & 44.28 & 44.34 & \textbf{45.07} & 45.04 & 44.96\\
    $3\times 4$            & SSIM & 0.997 & 0.995 & 0.997 & \textbf{0.999} & 0.998\\
    \hline
    Phone                  & PSNR & 36.98 & 37.34 & 37.65 & 37.72 & \textbf{38.15}\\
    $3\times 4$            & SSIM & 0.981 & 0.982 & 0.983 & 0.985 & \textbf{0.988}\\
    \hline
\end{tabular}
\end{table}
%%%%%%%%%%%%%%%%%%%%%%%%%%%%%%%%%%%%%%%%%%%%%%%%%%%%%%%%%%%%%%%%%%%%%%%%%%%%%
%%%%%%%%%%%%%%%%%%%%%%%%%%%%%%%%%%%%%%%%%%%%%%%%%%%%%%%%%%%%%%%%%%%%%%%%%%%%%
Table \ref{tab:synthetic contrast} shows the quantitive comparison. It is obvious that our network architecture outperforms the state-of-the-art methods on synthetic data. We attribute this to the additional FOV input and the deformable ResBlock, which endow the proposed architecture with spatial adaptability. Moreover, we retrain the state-of-the-art methods by uniformly separating all the data into $3 \times 4$ patches (overlapping with each other). The quantitive results are shown in Table \ref{tab:synthetic contrast}. We note that the model-based optimization methods make a leap in PSNR and SSIM when data are partitioned. For DLSR datasets, the PSFs are relatively consistent, so the IRCNN and Self-Deblur can achieve better quantitative results for patches with relatively consistent blur. For Phone datasets, the PSFs show a wide range of spatial variation over regions, which means that higher restoration cannot be achieved by simply partitioning the data. The smaller size of the patch might improve the restored result, but so would the computation overhead. In summary, our proposed architecture efficiently integrates the spatial information and will not occupy many more computing resources.

\subsubsection{Real Testing Image}
\label{sec:Real-Shoot Test Image}
%%%%%%%%%%%%%%%%%%%%%%%%%%%%%%%%%%%%%%%%%%%%%%%%%%%%%%%%%%%%%%%%%%%%%%%%%%%%%
%%%%%%%%%%%%%%%%%%%%%%%%%%%%%%%%%%%%%%%%%%%%%%%%%%%%%%%%%%%%%%%%%%%%%%%%%%%%%
\begin{figure*}[ht]
  \centering
  \includegraphics[width=\linewidth]{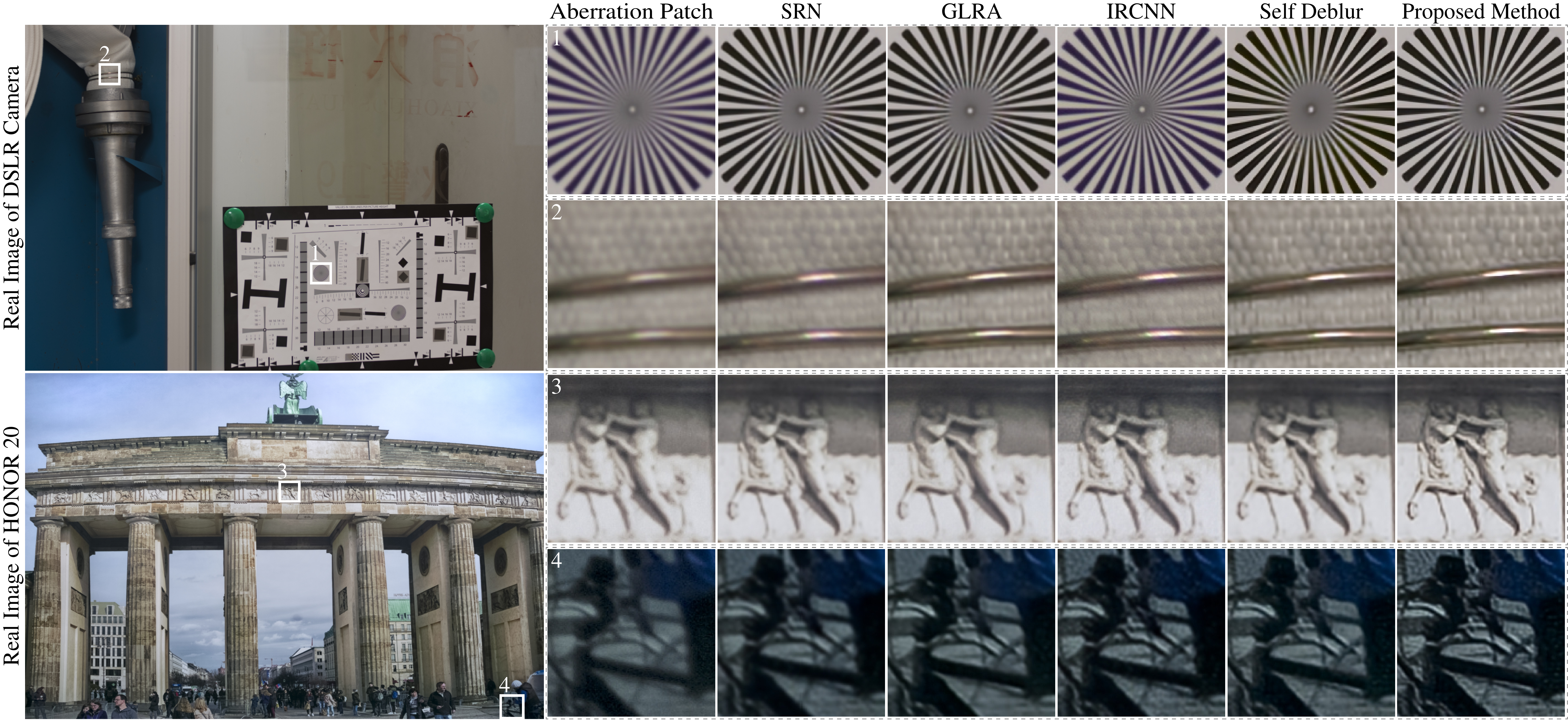}
  \caption{\textbf{Real-shoot image restoration comparison.} The first row and the second row present the test results of the DSLR camera lens. The third and the fourth row are the results of HUAWEI HONOR 20. In the first column, we highlight the corresponding regions of aberration patches with white boxes.}
  \Description{Real-shoot image restoration comparison.}
  \label{fig:Real-shoot image restoration comparison}
\end{figure*}
%%%%%%%%%%%%%%%%%%%%%%%%%%%%%%%%%%%%%%%%%%%%%%%%%%%%%%%%%%%%%%%%%%%%%%%%%%%%%
%%%%%%%%%%%%%%%%%%%%%%%%%%%%%%%%%%%%%%%%%%%%%%%%%%%%%%%%%%%%%%%%%%%%%%%%%%%%%
With the testing images taken by the camera (detailed in Sec.\ref{sec:Generate Testing Images}), we evaluate the proposed technique with state-of-the-art deblurring methods. As shown in Figure \ref{fig:Real-shoot image restoration comparison}, we show two patches of each test image: one patch is near the center of the image, and the other patch is on the edge. We note that the PSFs of the DSLR camera are relatively consistent (as shown in Figure \ref{fig:PSF characteristic}), so for the mainstream deblurring algorithms, which are used to deal with globally consistent blur, the restoration of the center and edge of the comparison methods are similar. When compared with other techniques, the results of the proposed approach in the central FOV are similar to those of other methods. In the edge of the image, our result of the DSLR camera is slightly better than those of the other methods. However, for the situation in which the PSFs show a wide range of spatial variation over regions, such as mobile phone cameras, the advantages of our approach are evident. In Figure \ref{fig:Real-shoot image restoration comparison}, one can see that the proposed method is superior to all other methods for center and edge restoration. The reconstruction effectively removes the blurring caused by optical aberrations, while the spatial-adaptive network architecture successfully handles the spatially variant PSFs and preserves the texture of the original image. We note that some results of GLRA and IRCNN are somewhat comparable to the proposed method, while in general, our approach can obtain better results in different FOVs. It must be emphasized that the additional input is useful for optical aberration correction because the degradation is highly correlated with the FOV information. Overall, we demonstrate the spatial adaptability of our network architecture. When compared with other methods, the proposed approach is more effective in dealing with the PSFs of spatial variation and therefore is suitable for optical aberration correction. 

\section{Experimental Assessment}
\label{sec:Experimental Assessment}
In this section, we take a comprehensive look at the proposed approach and demonstrate its potential applications. First, we assess how much the proposed method improves MTF, chromatismm and line-pair resolution. Second, the restored results are compared with the image postprocessed by HUAWEI ISP, to evaluate the improvement in integrating aberration correction into an ISP system.

\subsection{MTF and LPI Enhancement}
\label{sec:MTF and LPI Enhancement}
%%%%%%%%%%%%%%%%%%%%%%%%%%%%%%%%%%%%%%%%%%%%%%%%%%%%%%%%%%%%%%%%%%%%%%%%%%%%%
%%%%%%%%%%%%%%%%%%%%%%%%%%%%%%%%%%%%%%%%%%%%%%%%%%%%%%%%%%%%%%%%%%%%%%%%%%%%%
\begin{figure*}[ht]
  \centering
  \includegraphics[width=\linewidth]{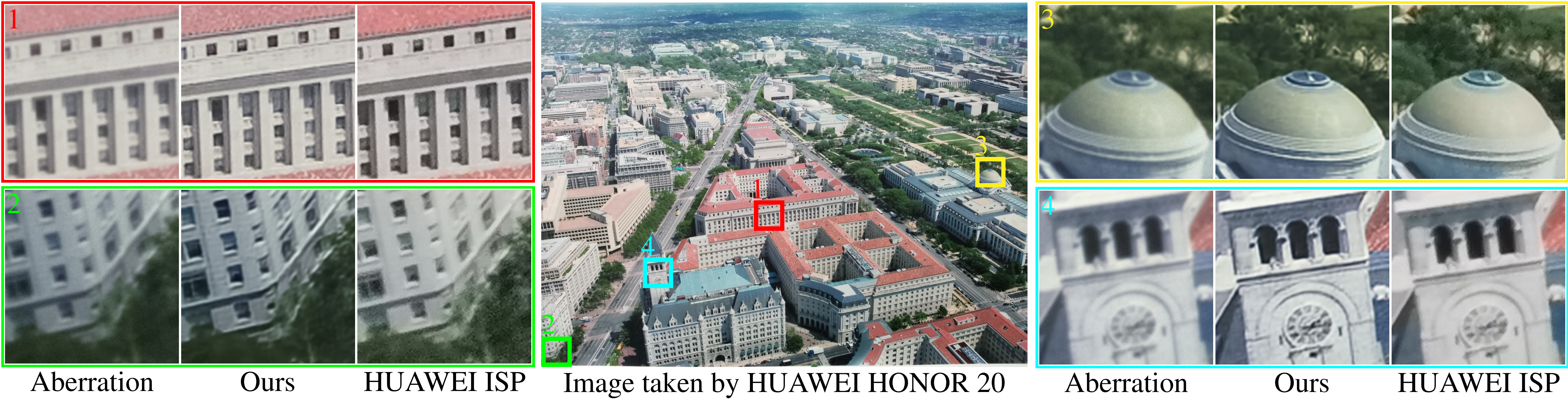}
  \caption{\textbf{Aberrations correction \textit{vs.} HUAWEI ISP.} Experimental results of the aberration correction pipeline and the HUAWEI ISP are amplified on both sides of the image. We show different FOVs' patches for validation.}
  \Description{Aberrations correction vs. HUAWEI ISP.}
  \label{fig:Aberrations correction vs. HUAWEI ISP}
\end{figure*}
%%%%%%%%%%%%%%%%%%%%%%%%%%%%%%%%%%%%%%%%%%%%%%%%%%%%%%%%%%%%%%%%%%%%%%%%%%%%%
%%%%%%%%%%%%%%%%%%%%%%%%%%%%%%%%%%%%%%%%%%%%%%%%%%%%%%%%%%%%%%%%%%%%%%%%%%%%%

%%%%%%%%%%%%%%%%%%%%%%%%%%%%%%%%%%%%%%%%%%%%%%%%%%%%%%%%%%%%%%%%%%%%%%%%%%%%%
%%%%%%%%%%%%%%%%%%%%%%%%%%%%%%%%%%%%%%%%%%%%%%%%%%%%%%%%%%%%%%%%%%%%%%%%%%%%%
\begin{figure}[t]
  \centering
  \includegraphics[width=\linewidth]{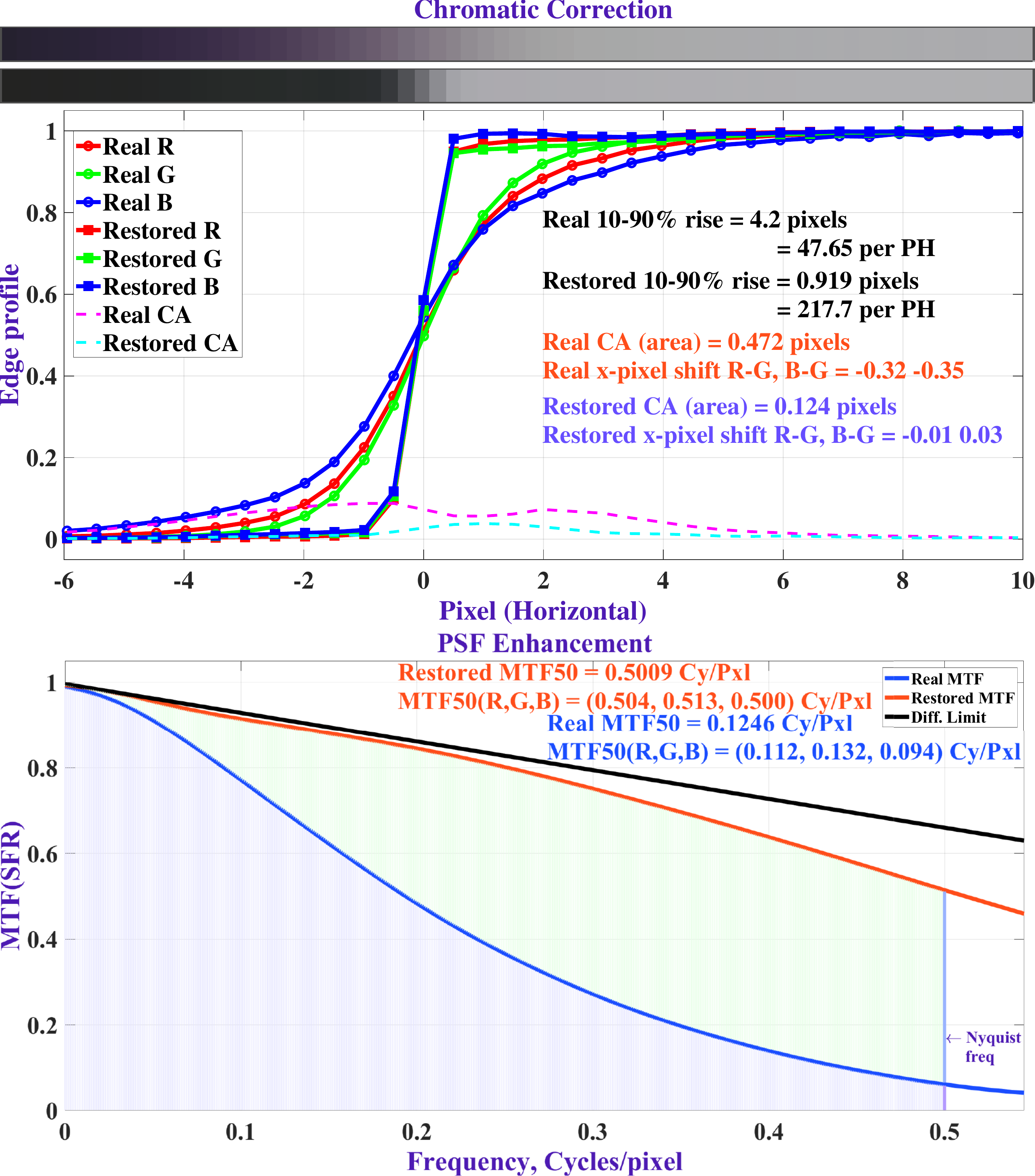}
  \caption{\textbf{Chromatic correction and MTF enhancement.} The detected edge of the degraded image and the restored edge are plotted on the top. The RGB chromatic aberrations, the average MTF of the real image and reconstruction, and the diffraction limitation MTF are sequentially plotted below. We also show chromatic aberration curves to evaluate the dispersion of RGB channels.}
  \Description{chro&mtf enhancement.}
  \label{fig:chro&mtf enhancement}
\end{figure}
%%%%%%%%%%%%%%%%%%%%%%%%%%%%%%%%%%%%%%%%%%%%%%%%%%%%%%%%%%%%%%%%%%%%%%%%%%%%%
%%%%%%%%%%%%%%%%%%%%%%%%%%%%%%%%%%%%%%%%%%%%%%%%%%%%%%%%%%%%%%%%%%%%%%%%%%%%%
We compare the chromatism and the MTF of degraded image and recovered image (shown in Figure \ref{fig:chro&mtf enhancement}). Though only trained with synthetic data, the proposed method can restore perfect knife edge and greatly improve the MTF results. Moreover, the pixel values of the R, G, and B channels of the reconstructed image are closer to each other, which indicates that our method successfully handles the chromatism between different channels. The lower restored CA curve also proves that the shift between RGB channels is smaller. Moreover, the enhancement of line-pair per inch (LPI) is shown in Figure \ref{fig:Real-shoot image restoration comparison}, where the line-pair resolution is improved in the center of the ISO12233 chart. We note that the restored MTF results approach the diffraction limit of the lens, which is the physical limit resolution of an imaging device.

\subsection{Aberrations correction \textit{vs.} HUAWEI ISP}
\label{sec:aberrations correction vs. HUAWEI ISP}

The ultimate goal of the proposed work is to integrate the module of aberration correction into an ISP system and to produce nearly perfect images. In today's handcrafted ISPs, sharpening is the core method for alleviating the blurring caused by optical aberrations. Unfortunately, the sharpening operation is globally consistent and cannot solve the spatially variant blur. Therefore, it introduces unsolved blur in the boundary of FOV and "ringing" effect in the center of figure (as shown in Figure \ref{fig:aberration ISP}c). In Figure \ref{fig:Aberrations correction vs. HUAWEI ISP}, we compare the results of our proposed pipeline (as shown in Figure \ref{fig:Overview}) and the images generated by HUAWEI ISP (\textbf{for more results, please refer to the supplementary materials}). Obviously, our proposed method adapts to the spatially variant degradation and the results are visually superior to HUAWEI's ISP. Together with accurate synthetic data and the customized restoration algorithm, we demonstrate that it is possible to handle optical aberrations in postprocessing. This also encourages us to use a lightweight lens for high-quality imaging.

\section{Discussion And Conclusion}
In this paper, we demonstrated that it is feasible to correct optical aberrations in postprocessing systems. Based on the optical PSF model, the imaging simulation system was developed. Moreover, with synthetic data generated in simulation, we designed the spatial-adaptive network for effective aberration correction.

To be more specific, we constructed an imaging simulation system based on the optical PSF model and the ISP system in the digital camera. Although the lenses in the real world are ever-changing, accurate imaging results can be obtained if crucial parameters are given. In this way, we established the datasets for a specific camera lens. Moreover, to correct the aberrations of optical design, we proposed a spatial-adaptive CNN that combines FOV information to handle aberrations of different sizes and shapes. After that, trained with the simulation datasets, the learned network model was used to map the real degraded images to the restored results. We address spatially variant blurring and chromatism by introducing FOV information, deformable ResBlock, and context block. The proposed method achieves perfect image quality and is more effective in dealing with PSFs varying with space when compared with state-of-the-art deblur methods. Moreover, we successfully corrected the aberrations of the DSLR camera and HUAWEI HONOR 20, which proves that it is a practical method for correcting aberrations and easily transport them to other lenses. Finally, we demonstrate that it is feasible to integrate aberration correction with the existing ISP systems, which will bring great positive benefits.

We find that the introduction of optical information and spatial adaptive capability can result in better recovery, but today's devices still can't afford the computational overhead of our postprocessing system. The dedicated NPUs and AI chips equipped on many modern imaging devices still limited to scene classification or postprocessing of a image with low resolution. Today's optical design and imaging postprocessing system are divisive and we strongly insist that the optical lens and ISP systems can be seen as encoder and decoder, respectively. In this way, the lens-ISP joint design is possible for modern imaging systems. Although this work focuses on the postprocessing system of cameras, we envision that it is practical to jointly design optical lenses and ISP systems for aberration correction or many other tasks such as 3D visual reconstruction. 

%%%%%%%%%%%%%%%%%%%%%%%%%%%%%%%%%%%%%%%%%%%%%%%%%%%%%%%%%%%%%%%%%%%%%%%%%%%%%
%%%%%%%%%%%%%%%%%%%%%%%%%%%%%%%%%%%%%%%%%%%%%%%%%%%%%%%%%%%%%%%%%%%%%%%%%%%%%

\begin{acks}
This work was supported by National Natural Science Foundation of China (NSFC) under Award No.61975175. We also thank Huawei for their support. Finally we would like to thank the TOG reviewers for their valuable comments.
\end{acks}

\bibliographystyle{ACM-Reference-Format}
\bibliography{main}


\begin{thebibliography}{71}

%%% the Web address.

\ifx \showCODEN    \undefined \def \showCODEN     #1{\unskip}     \fi
\ifx \showDOI      \undefined \def \showDOI       #1{#1}\fi
\ifx \showISBNx    \undefined \def \showISBNx     #1{\unskip}     \fi
\ifx \showISBNxiii \undefined \def \showISBNxiii  #1{\unskip}     \fi
\ifx \showISSN     \undefined \def \showISSN      #1{\unskip}     \fi
\ifx \showLCCN     \undefined \def \showLCCN      #1{\unskip}     \fi
\ifx \shownote     \undefined \def \shownote      #1{#1}          \fi
\ifx \showarticletitle \undefined \def \showarticletitle #1{#1}   \fi
\ifx \showURL      \undefined \def \showURL       {\relax}        \fi
% invisible to TeX
\providecommand\bibfield[2]{#2}
\providecommand\bibinfo[2]{#2}
\providecommand\natexlab[1]{#1}
\providecommand\showeprint[2][]{arXiv:#2}

\bibitem[\protect\citeauthoryear{Abdelhamed, Lin, and Brown}{Abdelhamed
  et~al\mbox{.}}{2018}]%
        {2018A}
\bibfield{author}{\bibinfo{person}{Abdelrahman Abdelhamed},
  \bibinfo{person}{Stephen Lin}, {and} \bibinfo{person}{Michael~S. Brown}.}
  \bibinfo{year}{2018}\natexlab{}.
\newblock \showarticletitle{A High-Quality Denoising Dataset for Smartphone
  Cameras}. In \bibinfo{booktitle}{\emph{Proceedings of the IEEE Conference on
  Computer Vision and Pattern Recognition (CVPR)}}. \bibinfo{publisher}{IEEE
  Computer Society}, \bibinfo{address}{Salt Lake City},
  \bibinfo{pages}{694--711}.
\newblock


\bibitem[\protect\citeauthoryear{Agustsson and Timofte}{Agustsson and
  Timofte}{2017}]%
        {agustsson2017ntire}
\bibfield{author}{\bibinfo{person}{Eirikur Agustsson} {and}
  \bibinfo{person}{Radu Timofte}.} \bibinfo{year}{2017}\natexlab{}.
\newblock \showarticletitle{NTIRE 2017 Challenge on Single Image
  Super-Resolution: Dataset and Study}. In
  \bibinfo{booktitle}{\emph{Proceedings of the IEEE Conference on Computer
  Vision and Pattern Recognition (CVPR) Workshops}}. \bibinfo{publisher}{IEEE
  Computer Society}, \bibinfo{address}{Hawaii}, \bibinfo{pages}{126--135}.
\newblock


\bibitem[\protect\citeauthoryear{Ahi}{Ahi}{2017}]%
        {ahi2017mathematical}
\bibfield{author}{\bibinfo{person}{Kiarash Ahi}.}
  \bibinfo{year}{2017}\natexlab{}.
\newblock \showarticletitle{Mathematical modeling of THz point spread function
  and simulation of THz imaging systems}.
\newblock \bibinfo{journal}{\emph{IEEE Transactions on Terahertz Science and
  Technology}} \bibinfo{volume}{7}, \bibinfo{number}{6} (\bibinfo{year}{2017}),
  \bibinfo{pages}{747--754}.
\newblock
\urldef\tempurl%
\url{https://doi.org/10.1109/TTHZ.2017.2750690}
\showDOI{\tempurl}


\bibitem[\protect\citeauthoryear{Baker and Copson}{Baker and Copson}{2003}]%
        {baker2003mathematical}
\bibfield{author}{\bibinfo{person}{Bevan~B Baker} {and}
  \bibinfo{person}{Edward~Thomas Copson}.} \bibinfo{year}{2003}\natexlab{}.
\newblock \bibinfo{booktitle}{\emph{The mathematical theory of Huygens'
  principle}}. Vol.~\bibinfo{volume}{329}.
\newblock \bibinfo{publisher}{American Mathematical Soc.},
  \bibinfo{address}{Boston}.
\newblock


\bibitem[\protect\citeauthoryear{Born and Wolf}{Born and Wolf}{2013}]%
        {born2013principles}
\bibfield{author}{\bibinfo{person}{Max Born} {and} \bibinfo{person}{Emil
  Wolf}.} \bibinfo{year}{2013}\natexlab{}.
\newblock \bibinfo{booktitle}{\emph{Principles of optics: electromagnetic
  theory of propagation, interference and diffraction of light}}.
\newblock \bibinfo{publisher}{Elsevier}, \bibinfo{address}{Boston}.
\newblock


\bibitem[\protect\citeauthoryear{Brooks, Mildenhall, Xue, Chen, Sharlet, and
  Barron}{Brooks et~al\mbox{.}}{2019}]%
        {brooks2019unprocessing}
\bibfield{author}{\bibinfo{person}{Tim Brooks}, \bibinfo{person}{Ben
  Mildenhall}, \bibinfo{person}{Tianfan Xue}, \bibinfo{person}{Jiawen Chen},
  \bibinfo{person}{Dillon Sharlet}, {and} \bibinfo{person}{Jonathan~T Barron}.}
  \bibinfo{year}{2019}\natexlab{}.
\newblock \showarticletitle{Unprocessing images for learned raw denoising}. In
  \bibinfo{booktitle}{\emph{Proceedings of the IEEE Conference on Computer
  Vision and Pattern Recognition}}. \bibinfo{publisher}{IEEE Computer Soc.},
  \bibinfo{address}{Columbus}, \bibinfo{pages}{11036--11045}.
\newblock


\bibitem[\protect\citeauthoryear{Buades, Coll, and Morel}{Buades
  et~al\mbox{.}}{2005}]%
        {buades2005non}
\bibfield{author}{\bibinfo{person}{Antoni Buades}, \bibinfo{person}{Bartomeu
  Coll}, {and} \bibinfo{person}{J-M Morel}.} \bibinfo{year}{2005}\natexlab{}.
\newblock \showarticletitle{A non-local algorithm for image denoising}. In
  \bibinfo{booktitle}{\emph{2005 IEEE Computer Society Conference on Computer
  Vision and Pattern Recognition (CVPR'05)}}, Vol.~\bibinfo{volume}{2}. IEEE,
  \bibinfo{publisher}{IEEE}, \bibinfo{address}{Portland},
  \bibinfo{pages}{60--65}.
\newblock


\bibitem[\protect\citeauthoryear{Cai, Zeng, Yong, Cao, and Zhang}{Cai
  et~al\mbox{.}}{2020}]%
        {2020Toward}
\bibfield{author}{\bibinfo{person}{J. Cai}, \bibinfo{person}{H. Zeng},
  \bibinfo{person}{H. Yong}, \bibinfo{person}{Z. Cao}, {and}
  \bibinfo{person}{L. Zhang}.} \bibinfo{year}{2020}\natexlab{}.
\newblock \showarticletitle{Toward Real-World Single Image Super-Resolution: A
  New Benchmark and a New Model}. In \bibinfo{booktitle}{\emph{2019 IEEE/CVF
  International Conference on Computer Vision (ICCV)}}.
  \bibinfo{publisher}{IEEE Computer Soc.}, \bibinfo{address}{Munich},
  \bibinfo{pages}{15--463}.
\newblock


\bibitem[\protect\citeauthoryear{Chang, Li, Feng, and Xu}{Chang
  et~al\mbox{.}}{2020}]%
        {chang2020spatial}
\bibfield{author}{\bibinfo{person}{Meng Chang}, \bibinfo{person}{Qi Li},
  \bibinfo{person}{Huajun Feng}, {and} \bibinfo{person}{Zhihai Xu}.}
  \bibinfo{year}{2020}\natexlab{}.
\newblock \showarticletitle{Spatial-Adaptive Network for Single Image
  Denoising}.
\newblock \bibinfo{journal}{\emph{arXiv preprint arXiv:2001.10291}}
  \bibinfo{volume}{26}, \bibinfo{number}{10} (\bibinfo{year}{2020}),
  \bibinfo{pages}{171--187}.
\newblock


\bibitem[\protect\citeauthoryear{Chen, Chen, Xu, and Koltun}{Chen
  et~al\mbox{.}}{2018}]%
        {chen2018learning}
\bibfield{author}{\bibinfo{person}{Chen Chen}, \bibinfo{person}{Qifeng Chen},
  \bibinfo{person}{Jia Xu}, {and} \bibinfo{person}{Vladlen Koltun}.}
  \bibinfo{year}{2018}\natexlab{}.
\newblock \showarticletitle{Learning to see in the dark}. In
  \bibinfo{booktitle}{\emph{Proceedings of the IEEE Conference on Computer
  Vision and Pattern Recognition}}. \bibinfo{publisher}{IEEE},
  \bibinfo{address}{Venice}, \bibinfo{pages}{3291--3300}.
\newblock


\bibitem[\protect\citeauthoryear{Condat}{Condat}{2010}]%
        {condat2010simple}
\bibfield{author}{\bibinfo{person}{Laurent Condat}.}
  \bibinfo{year}{2010}\natexlab{}.
\newblock \showarticletitle{A simple, fast and efficient approach to
  denoisaicking: Joint demosaicking and denoising}. In
  \bibinfo{booktitle}{\emph{2010 IEEE International Conference on Image
  Processing}}. IEEE, \bibinfo{publisher}{IEEE}, \bibinfo{address}{IEEE},
  \bibinfo{pages}{905--908}.
\newblock


\bibitem[\protect\citeauthoryear{Dai, Qi, Xiong, Li, Zhang, Hu, and Wei}{Dai
  et~al\mbox{.}}{2017}]%
        {dai2017deformable}
\bibfield{author}{\bibinfo{person}{Jifeng Dai}, \bibinfo{person}{Haozhi Qi},
  \bibinfo{person}{Yuwen Xiong}, \bibinfo{person}{Yi Li},
  \bibinfo{person}{Guodong Zhang}, \bibinfo{person}{Han Hu}, {and}
  \bibinfo{person}{Yichen Wei}.} \bibinfo{year}{2017}\natexlab{}.
\newblock \showarticletitle{Deformable convolutional networks}. In
  \bibinfo{booktitle}{\emph{Proceedings of the IEEE international conference on
  computer vision}}. \bibinfo{publisher}{IEEE}, \bibinfo{address}{IEEE},
  \bibinfo{pages}{764--773}.
\newblock


\bibitem[\protect\citeauthoryear{Dong, Loy, He, and Tang}{Dong
  et~al\mbox{.}}{2016}]%
        {dong2015image}
\bibfield{author}{\bibinfo{person}{Chao Dong}, \bibinfo{person}{Chen~Change
  Loy}, \bibinfo{person}{Kaiming He}, {and} \bibinfo{person}{Xiaoou Tang}.}
  \bibinfo{year}{2016}\natexlab{}.
\newblock \showarticletitle{Image Super-Resolution Using Deep Convolutional
  Networks}.
\newblock \bibinfo{journal}{\emph{IEEE Transactions on Pattern Analysis and
  Machine Intelligence}} \bibinfo{volume}{38}, \bibinfo{number}{2}
  (\bibinfo{year}{2016}), \bibinfo{pages}{295--307}.
\newblock
\urldef\tempurl%
\url{https://doi.org/10.1109/TPAMI.2015.2439281}
\showDOI{\tempurl}


\bibitem[\protect\citeauthoryear{Dubois}{Dubois}{2006}]%
        {dubois2006filter}
\bibfield{author}{\bibinfo{person}{Eric Dubois}.}
  \bibinfo{year}{2006}\natexlab{}.
\newblock \showarticletitle{Filter design for adaptive frequency-domain Bayer
  demosaicking}. In \bibinfo{booktitle}{\emph{2006 International Conference on
  Image Processing}}. IEEE, \bibinfo{publisher}{IEEE}, \bibinfo{address}{IEEE},
  \bibinfo{pages}{2705--2708}.
\newblock


\bibitem[\protect\citeauthoryear{Foi, Trimeche, Katkovnik, and Egiazarian}{Foi
  et~al\mbox{.}}{2008}]%
        {foi2008practical}
\bibfield{author}{\bibinfo{person}{Alessandro Foi}, \bibinfo{person}{Mejdi
  Trimeche}, \bibinfo{person}{Vladimir Katkovnik}, {and} \bibinfo{person}{Karen
  Egiazarian}.} \bibinfo{year}{2008}\natexlab{}.
\newblock \showarticletitle{Practical Poissonian-Gaussian Noise Modeling and
  Fitting for Single-Image Raw-Data}.
\newblock \bibinfo{journal}{\emph{IEEE Transactions on Image Processing}}
  \bibinfo{volume}{17}, \bibinfo{number}{10} (\bibinfo{year}{2008}),
  \bibinfo{pages}{1737--1754}.
\newblock
\urldef\tempurl%
\url{https://doi.org/10.1109/TIP.2008.2001399}
\showDOI{\tempurl}


\bibitem[\protect\citeauthoryear{Gijsenij, Gevers, and van~de Weijer}{Gijsenij
  et~al\mbox{.}}{2012}]%
        {gijsenij2011improving}
\bibfield{author}{\bibinfo{person}{Arjan Gijsenij}, \bibinfo{person}{Theo
  Gevers}, {and} \bibinfo{person}{Joost van~de Weijer}.}
  \bibinfo{year}{2012}\natexlab{}.
\newblock \showarticletitle{Improving Color Constancy by Photometric Edge
  Weighting}.
\newblock \bibinfo{journal}{\emph{IEEE Transactions on Pattern Analysis and
  Machine Intelligence}} \bibinfo{volume}{34}, \bibinfo{number}{5}
  (\bibinfo{year}{2012}), \bibinfo{pages}{918--929}.
\newblock
\urldef\tempurl%
\url{https://doi.org/10.1109/TPAMI.2011.197}
\showDOI{\tempurl}


\bibitem[\protect\citeauthoryear{Hasinoff, Sharlet, Geiss, Adams, Barron,
  Kainz, Chen, and Levoy}{Hasinoff et~al\mbox{.}}{2016}]%
        {hasinoff2016burst}
\bibfield{author}{\bibinfo{person}{Samuel~W. Hasinoff}, \bibinfo{person}{Dillon
  Sharlet}, \bibinfo{person}{Ryan Geiss}, \bibinfo{person}{Andrew Adams},
  \bibinfo{person}{Jonathan~T. Barron}, \bibinfo{person}{Florian Kainz},
  \bibinfo{person}{Jiawen Chen}, {and} \bibinfo{person}{Marc Levoy}.}
  \bibinfo{year}{2016}\natexlab{}.
\newblock \showarticletitle{Burst Photography for High Dynamic Range and
  Low-Light Imaging on Mobile Cameras}.
\newblock \bibinfo{journal}{\emph{ACM Trans. Graph.}} \bibinfo{volume}{35},
  \bibinfo{number}{6}, Article \bibinfo{articleno}{192} (\bibinfo{date}{nov}
  \bibinfo{year}{2016}), \bibinfo{numpages}{12}~pages.
\newblock
\showISSN{0730-0301}
\urldef\tempurl%
\url{https://doi.org/10.1145/2980179.2980254}
\showDOI{\tempurl}


\bibitem[\protect\citeauthoryear{Heckbert}{Heckbert}{1995}]%
        {heckbert1995fourier}
\bibfield{author}{\bibinfo{person}{Paul Heckbert}.}
  \bibinfo{year}{1995}\natexlab{}.
\newblock \showarticletitle{Fourier transforms and the fast Fourier transform
  (FFT) algorithm}.
\newblock \bibinfo{journal}{\emph{Computer Graphics}}  \bibinfo{volume}{2}
  (\bibinfo{year}{1995}), \bibinfo{pages}{15--463}.
\newblock


\bibitem[\protect\citeauthoryear{Heide, Rouf, Hullin, Labitzke, Heidrich, and
  Kolb}{Heide et~al\mbox{.}}{2013}]%
        {heide2013high}
\bibfield{author}{\bibinfo{person}{Felix Heide}, \bibinfo{person}{Mushfiqur
  Rouf}, \bibinfo{person}{Matthias~B. Hullin}, \bibinfo{person}{Bjorn
  Labitzke}, \bibinfo{person}{Wolfgang Heidrich}, {and}
  \bibinfo{person}{Andreas Kolb}.} \bibinfo{year}{2013}\natexlab{}.
\newblock \showarticletitle{High-Quality Computational Imaging through Simple
  Lenses}.
\newblock \bibinfo{journal}{\emph{ACM Trans. Graph.}} \bibinfo{volume}{32},
  \bibinfo{number}{5}, Article \bibinfo{articleno}{149} (\bibinfo{date}{oct}
  \bibinfo{year}{2013}), \bibinfo{numpages}{14}~pages.
\newblock
\showISSN{0730-0301}
\urldef\tempurl%
\url{https://doi.org/10.1145/2516971.2516974}
\showDOI{\tempurl}


\bibitem[\protect\citeauthoryear{Heide, Steinberger, Tsai, Rouf, Paj\k{a}k,
  Reddy, Gallo, Liu, Heidrich, Egiazarian, Kautz, and Pulli}{Heide
  et~al\mbox{.}}{2014}]%
        {heide2014flexisp}
\bibfield{author}{\bibinfo{person}{Felix Heide}, \bibinfo{person}{Markus
  Steinberger}, \bibinfo{person}{Yun-Ta Tsai}, \bibinfo{person}{Mushfiqur
  Rouf}, \bibinfo{person}{Dawid Paj\k{a}k}, \bibinfo{person}{Dikpal Reddy},
  \bibinfo{person}{Orazio Gallo}, \bibinfo{person}{Jing Liu},
  \bibinfo{person}{Wolfgang Heidrich}, \bibinfo{person}{Karen Egiazarian},
  \bibinfo{person}{Jan Kautz}, {and} \bibinfo{person}{Kari Pulli}.}
  \bibinfo{year}{2014}\natexlab{}.
\newblock \showarticletitle{FlexISP: A Flexible Camera Image Processing
  Framework}.
\newblock \bibinfo{journal}{\emph{ACM Trans. Graph.}} \bibinfo{volume}{33},
  \bibinfo{number}{6}, Article \bibinfo{articleno}{231} (\bibinfo{date}{Nov.}
  \bibinfo{year}{2014}), \bibinfo{numpages}{13}~pages.
\newblock
\showISSN{0730-0301}
\urldef\tempurl%
\url{https://doi.org/10.1145/2661229.2661260}
\showDOI{\tempurl}


\bibitem[\protect\citeauthoryear{Herbel, Kacprzak, Amara, Refregier, and
  Lucchi}{Herbel et~al\mbox{.}}{2018}]%
        {herbel2018fast}
\bibfield{author}{\bibinfo{person}{Jörg Herbel}, \bibinfo{person}{Tomasz
  Kacprzak}, \bibinfo{person}{Adam Amara}, \bibinfo{person}{Alexandre
  Refregier}, {and} \bibinfo{person}{Aurelien Lucchi}.}
  \bibinfo{year}{2018}\natexlab{}.
\newblock \showarticletitle{Fast point spread function modeling with deep
  learning}.
\newblock \bibinfo{journal}{\emph{Journal of Cosmology and Astroparticle
  Physics}} \bibinfo{volume}{2018}, \bibinfo{number}{07} (\bibinfo{date}{jul}
  \bibinfo{year}{2018}), \bibinfo{pages}{054--054}.
\newblock
\urldef\tempurl%
\url{https://doi.org/10.1088/1475-7516/2018/07/054}
\showDOI{\tempurl}


\bibitem[\protect\citeauthoryear{Hirakawa and Parks}{Hirakawa and
  Parks}{2005}]%
        {hirakawa2005adaptive}
\bibfield{author}{\bibinfo{person}{K. Hirakawa} {and} \bibinfo{person}{T.W.
  Parks}.} \bibinfo{year}{2005}\natexlab{}.
\newblock \showarticletitle{Adaptive homogeneity-directed demosaicing
  algorithm}.
\newblock \bibinfo{journal}{\emph{IEEE Transactions on Image Processing}}
  \bibinfo{volume}{14}, \bibinfo{number}{3} (\bibinfo{year}{2005}),
  \bibinfo{pages}{360--369}.
\newblock
\urldef\tempurl%
\url{https://doi.org/10.1109/TIP.2004.838691}
\showDOI{\tempurl}


\bibitem[\protect\citeauthoryear{Hirsch and Scholkopf}{Hirsch and
  Scholkopf}{2015}]%
        {hirsch2015self}
\bibfield{author}{\bibinfo{person}{Michael Hirsch} {and}
  \bibinfo{person}{Bernhard Scholkopf}.} \bibinfo{year}{2015}\natexlab{}.
\newblock \showarticletitle{Self-calibration of optical lenses}. In
  \bibinfo{booktitle}{\emph{Proceedings of the IEEE International Conference on
  Computer Vision}}. \bibinfo{publisher}{IEEE}, \bibinfo{address}{IEEE},
  \bibinfo{pages}{612--620}.
\newblock


\bibitem[\protect\citeauthoryear{Ignatov, Kobyshev, Timofte, Vanhoey, and
  Van~Gool}{Ignatov et~al\mbox{.}}{2017}]%
        {ignatov2017dslr}
\bibfield{author}{\bibinfo{person}{Andrey Ignatov}, \bibinfo{person}{Nikolay
  Kobyshev}, \bibinfo{person}{Radu Timofte}, \bibinfo{person}{Kenneth Vanhoey},
  {and} \bibinfo{person}{Luc Van~Gool}.} \bibinfo{year}{2017}\natexlab{}.
\newblock \showarticletitle{Dslr-quality photos on mobile devices with deep
  convolutional networks}. In \bibinfo{booktitle}{\emph{Proceedings of the IEEE
  International Conference on Computer Vision}}. \bibinfo{publisher}{IEEE},
  \bibinfo{address}{IEEE}, \bibinfo{pages}{3277--3285}.
\newblock


\bibitem[\protect\citeauthoryear{Ignatov, Kobyshev, Timofte, Vanhoey, and
  Van~Gool}{Ignatov et~al\mbox{.}}{2018}]%
        {ignatov2018wespe}
\bibfield{author}{\bibinfo{person}{Andrey Ignatov}, \bibinfo{person}{Nikolay
  Kobyshev}, \bibinfo{person}{Radu Timofte}, \bibinfo{person}{Kenneth Vanhoey},
  {and} \bibinfo{person}{Luc Van~Gool}.} \bibinfo{year}{2018}\natexlab{}.
\newblock \showarticletitle{Wespe: weakly supervised photo enhancer for digital
  cameras}. In \bibinfo{booktitle}{\emph{Proceedings of the IEEE Conference on
  Computer Vision and Pattern Recognition Workshops}}.
  \bibinfo{publisher}{IEEE}, \bibinfo{address}{IEEE},
  \bibinfo{pages}{691--700}.
\newblock


\bibitem[\protect\citeauthoryear{Ignatov and Timofte}{Ignatov and
  Timofte}{2019}]%
        {ignatov2019ntire}
\bibfield{author}{\bibinfo{person}{Andrey Ignatov} {and} \bibinfo{person}{Radu
  Timofte}.} \bibinfo{year}{2019}\natexlab{}.
\newblock \showarticletitle{Ntire 2019 challenge on image enhancement: Methods
  and results}. In \bibinfo{booktitle}{\emph{Proceedings of the IEEE Conference
  on Computer Vision and Pattern Recognition Workshops}}.
  \bibinfo{publisher}{IEEE}, \bibinfo{address}{IEEE}, \bibinfo{pages}{0--0}.
\newblock


\bibitem[\protect\citeauthoryear{Ignatov, Van~Gool, and Timofte}{Ignatov
  et~al\mbox{.}}{2020}]%
        {ignatov2020replacing}
\bibfield{author}{\bibinfo{person}{Andrey Ignatov}, \bibinfo{person}{Luc
  Van~Gool}, {and} \bibinfo{person}{Radu Timofte}.}
  \bibinfo{year}{2020}\natexlab{}.
\newblock \showarticletitle{Replacing mobile camera isp with a single deep
  learning model}. In \bibinfo{booktitle}{\emph{Proceedings of the IEEE/CVF
  Conference on Computer Vision and Pattern Recognition Workshops}}.
  \bibinfo{publisher}{IEEE}, \bibinfo{address}{IEEE},
  \bibinfo{pages}{536--537}.
\newblock


\bibitem[\protect\citeauthoryear{Jemec, Pernu{\v{s}}, Likar, and
  B{\"u}rmen}{Jemec et~al\mbox{.}}{2017}]%
        {jemec20172d}
\bibfield{author}{\bibinfo{person}{Jurij Jemec}, \bibinfo{person}{Franjo
  Pernu{\v{s}}}, \bibinfo{person}{Bo{\v{s}}tjan Likar}, {and}
  \bibinfo{person}{Miran B{\"u}rmen}.} \bibinfo{year}{2017}\natexlab{}.
\newblock \showarticletitle{2D sub-pixel point spread function measurement
  using a virtual point-like source}.
\newblock \bibinfo{journal}{\emph{International journal of computer vision}}
  \bibinfo{volume}{121}, \bibinfo{number}{3} (\bibinfo{year}{2017}),
  \bibinfo{pages}{391--402}.
\newblock


\bibitem[\protect\citeauthoryear{Ji, Y, Y, Wang, Li, and Huang}{Ji
  et~al\mbox{.}}{2020}]%
        {2020Real}
\bibfield{author}{\bibinfo{person}{X. Ji}, \bibinfo{person}{Cao Y},
  \bibinfo{person}{Tai Y}, \bibinfo{person}{C. Wang}, \bibinfo{person}{J. Li},
  {and} \bibinfo{person}{F. Huang}.} \bibinfo{year}{2020}\natexlab{}.
\newblock \showarticletitle{Real-World Super-Resolution via Kernel Estimation
  and Noise Injection}. In \bibinfo{booktitle}{\emph{2020 IEEE/CVF Conference
  on Computer Vision and Pattern Recognition Workshops (CVPRW)}}.
  \bibinfo{publisher}{IEEE}, \bibinfo{address}{IEEE}, \bibinfo{pages}{15--463}.
\newblock


\bibitem[\protect\citeauthoryear{Johnson, Alahi, and Fei-Fei}{Johnson
  et~al\mbox{.}}{2016}]%
        {johnson2016perceptual}
\bibfield{author}{\bibinfo{person}{Justin Johnson}, \bibinfo{person}{Alexandre
  Alahi}, {and} \bibinfo{person}{Li Fei-Fei}.} \bibinfo{year}{2016}\natexlab{}.
\newblock \showarticletitle{Perceptual losses for real-time style transfer and
  super-resolution}. In \bibinfo{booktitle}{\emph{European conference on
  computer vision}}. Springer, \bibinfo{publisher}{Springer},
  \bibinfo{address}{Boston}, \bibinfo{pages}{694--711}.
\newblock


\bibitem[\protect\citeauthoryear{Kim, Kwon~Lee, and Mu~Lee}{Kim
  et~al\mbox{.}}{2016}]%
        {kim2016accurate}
\bibfield{author}{\bibinfo{person}{Jiwon Kim}, \bibinfo{person}{Jung Kwon~Lee},
  {and} \bibinfo{person}{Kyoung Mu~Lee}.} \bibinfo{year}{2016}\natexlab{}.
\newblock \showarticletitle{Accurate image super-resolution using very deep
  convolutional networks}. In \bibinfo{booktitle}{\emph{Proceedings of the IEEE
  conference on computer vision and pattern recognition}}.
  \bibinfo{publisher}{IEEE}, \bibinfo{address}{Venice},
  \bibinfo{pages}{1646--1654}.
\newblock


\bibitem[\protect\citeauthoryear{Kwok, Shi, Ha, Fang, Chen, and Jia}{Kwok
  et~al\mbox{.}}{2013}]%
        {kwok2013simultaneous}
\bibfield{author}{\bibinfo{person}{N.M. Kwok}, \bibinfo{person}{H.Y. Shi},
  \bibinfo{person}{Q.P. Ha}, \bibinfo{person}{G. Fang}, \bibinfo{person}{S.Y.
  Chen}, {and} \bibinfo{person}{X. Jia}.} \bibinfo{year}{2013}\natexlab{}.
\newblock \showarticletitle{Simultaneous image color correction and enhancement
  using particle swarm optimization}.
\newblock \bibinfo{journal}{\emph{Engineering Applications of Artificial
  Intelligence}} \bibinfo{volume}{26}, \bibinfo{number}{10}
  (\bibinfo{year}{2013}), \bibinfo{pages}{2356--2371}.
\newblock
\showISSN{0952-1976}
\urldef\tempurl%
\url{https://doi.org/10.1016/j.engappai.2013.07.023}
\showDOI{\tempurl}


\bibitem[\protect\citeauthoryear{Ledig, Theis, Husz{\'a}r, Caballero,
  Cunningham, Acosta, Aitken, Tejani, Totz, Wang, et~al\mbox{.}}{Ledig
  et~al\mbox{.}}{2017}]%
        {ledig2017photo}
\bibfield{author}{\bibinfo{person}{Christian Ledig}, \bibinfo{person}{Lucas
  Theis}, \bibinfo{person}{Ferenc Husz{\'a}r}, \bibinfo{person}{Jose
  Caballero}, \bibinfo{person}{Andrew Cunningham}, \bibinfo{person}{Alejandro
  Acosta}, \bibinfo{person}{Andrew Aitken}, \bibinfo{person}{Alykhan Tejani},
  \bibinfo{person}{Johannes Totz}, \bibinfo{person}{Zehan Wang},
  {et~al\mbox{.}}} \bibinfo{year}{2017}\natexlab{}.
\newblock \showarticletitle{Photo-realistic single image super-resolution using
  a generative adversarial network}. In \bibinfo{booktitle}{\emph{Proceedings
  of the IEEE conference on computer vision and pattern recognition}}.
  \bibinfo{publisher}{IEEE}, \bibinfo{address}{Venice},
  \bibinfo{pages}{4681--4690}.
\newblock


\bibitem[\protect\citeauthoryear{Li, Gunturk, and Zhang}{Li
  et~al\mbox{.}}{2008}]%
        {li2008image}
\bibfield{author}{\bibinfo{person}{Xin Li}, \bibinfo{person}{Bahadir Gunturk},
  {and} \bibinfo{person}{Lei Zhang}.} \bibinfo{year}{2008}\natexlab{}.
\newblock \showarticletitle{Image demosaicing: A systematic survey}. In
  \bibinfo{booktitle}{\emph{Visual Communications and Image Processing 2008}},
  Vol.~\bibinfo{volume}{6822}. International Society for Optics and Photonics,
  \bibinfo{publisher}{IEEE}, \bibinfo{address}{Venice},
  \bibinfo{pages}{68221J}.
\newblock


\bibitem[\protect\citeauthoryear{Lian, Chang, Tan, and Zagorodnov}{Lian
  et~al\mbox{.}}{2007}]%
        {lian2007adaptive}
\bibfield{author}{\bibinfo{person}{Nai-Xiang Lian}, \bibinfo{person}{Lanlan
  Chang}, \bibinfo{person}{Yap-Peng Tan}, {and} \bibinfo{person}{Vitali
  Zagorodnov}.} \bibinfo{year}{2007}\natexlab{}.
\newblock \showarticletitle{Adaptive Filtering for Color Filter Array
  Demosaicking}.
\newblock \bibinfo{journal}{\emph{IEEE Transactions on Image Processing}}
  \bibinfo{volume}{16}, \bibinfo{number}{10} (\bibinfo{year}{2007}),
  \bibinfo{pages}{2515--2525}.
\newblock
\urldef\tempurl%
\url{https://doi.org/10.1109/TIP.2007.904459}
\showDOI{\tempurl}


\bibitem[\protect\citeauthoryear{Mahajan}{Mahajan}{1982}]%
        {mahajan1982strehl}
\bibfield{author}{\bibinfo{person}{Virendra~N Mahajan}.}
  \bibinfo{year}{1982}\natexlab{}.
\newblock \showarticletitle{Strehl ratio for primary aberrations: some
  analytical results for circular and annular pupils}.
\newblock \bibinfo{journal}{\emph{JOSA}} \bibinfo{volume}{72},
  \bibinfo{number}{9} (\bibinfo{year}{1982}), \bibinfo{pages}{1258--1266}.
\newblock


\bibitem[\protect\citeauthoryear{Manual}{Manual}{2009}]%
        {manual2009optical}
\bibfield{author}{\bibinfo{person}{Zemax Manual}.}
  \bibinfo{year}{2009}\natexlab{}.
\newblock \bibinfo{title}{Optical Design Program User's Guide [R]}.
\newblock
\newblock


\bibitem[\protect\citeauthoryear{Martin, Fowlkes, Tal, and Malik}{Martin
  et~al\mbox{.}}{2001}]%
        {martin2001database}
\bibfield{author}{\bibinfo{person}{David Martin}, \bibinfo{person}{Charless
  Fowlkes}, \bibinfo{person}{Doron Tal}, {and} \bibinfo{person}{Jitendra
  Malik}.} \bibinfo{year}{2001}\natexlab{}.
\newblock \showarticletitle{A database of human segmented natural images and
  its application to evaluating segmentation algorithms and measuring
  ecological statistics}. In \bibinfo{booktitle}{\emph{Proceedings Eighth IEEE
  International Conference on Computer Vision. ICCV 2001}},
  Vol.~\bibinfo{volume}{2}. IEEE, \bibinfo{publisher}{IEEE},
  \bibinfo{address}{Venice}, \bibinfo{pages}{416--423}.
\newblock


\bibitem[\protect\citeauthoryear{Mei, Li, Zhang, Wu, Li, and Huang}{Mei
  et~al\mbox{.}}{2019}]%
        {mei2019higher}
\bibfield{author}{\bibinfo{person}{Kangfu Mei}, \bibinfo{person}{Juncheng Li},
  \bibinfo{person}{Jiajie Zhang}, \bibinfo{person}{Haoyu Wu},
  \bibinfo{person}{Jie Li}, {and} \bibinfo{person}{Rui Huang}.}
  \bibinfo{year}{2019}\natexlab{}.
\newblock \showarticletitle{Higher-resolution network for image demosaicing and
  enhancing}. In \bibinfo{booktitle}{\emph{2019 IEEE/CVF International
  Conference on Computer Vision Workshop (ICCVW)}}. IEEE,
  \bibinfo{publisher}{IEEE}, \bibinfo{address}{Venice},
  \bibinfo{pages}{3441--3448}.
\newblock


\bibitem[\protect\citeauthoryear{Mosleh, Green, Onzon, Begin, and
  Pierre~Langlois}{Mosleh et~al\mbox{.}}{2015}]%
        {mosleh2015camera}
\bibfield{author}{\bibinfo{person}{Ali Mosleh}, \bibinfo{person}{Paul Green},
  \bibinfo{person}{Emmanuel Onzon}, \bibinfo{person}{Isabelle Begin}, {and}
  \bibinfo{person}{JM Pierre~Langlois}.} \bibinfo{year}{2015}\natexlab{}.
\newblock \showarticletitle{Camera intrinsic blur kernel estimation: A reliable
  framework}. In \bibinfo{booktitle}{\emph{Proceedings of the IEEE conference
  on computer vision and pattern recognition}}. \bibinfo{publisher}{IEEE},
  \bibinfo{address}{Venice}, \bibinfo{pages}{4961--4968}.
\newblock


\bibitem[\protect\citeauthoryear{Nah, Hyun~Kim, and Mu~Lee}{Nah
  et~al\mbox{.}}{2017}]%
        {nah2017deep}
\bibfield{author}{\bibinfo{person}{Seungjun Nah}, \bibinfo{person}{Tae
  Hyun~Kim}, {and} \bibinfo{person}{Kyoung Mu~Lee}.}
  \bibinfo{year}{2017}\natexlab{}.
\newblock \showarticletitle{Deep multi-scale convolutional neural network for
  dynamic scene deblurring}. In \bibinfo{booktitle}{\emph{Proceedings of the
  IEEE Conference on Computer Vision and Pattern Recognition}}.
  \bibinfo{publisher}{IEEE}, \bibinfo{address}{Venice},
  \bibinfo{pages}{3883--3891}.
\newblock


\bibitem[\protect\citeauthoryear{Pan, Sun, Pfister, and Yang}{Pan
  et~al\mbox{.}}{2016}]%
        {pan2016blind}
\bibfield{author}{\bibinfo{person}{Jinshan Pan}, \bibinfo{person}{Deqing Sun},
  \bibinfo{person}{Hanspeter Pfister}, {and} \bibinfo{person}{Ming-Hsuan
  Yang}.} \bibinfo{year}{2016}\natexlab{}.
\newblock \showarticletitle{Blind image deblurring using dark channel prior}.
  In \bibinfo{booktitle}{\emph{Proceedings of the IEEE Conference on Computer
  Vision and Pattern Recognition}}. \bibinfo{publisher}{IEEE},
  \bibinfo{address}{Venice}, \bibinfo{pages}{1628--1636}.
\newblock


\bibitem[\protect\citeauthoryear{Paul}{Paul}{2014}]%
        {paul2014huygens}
\bibfield{author}{\bibinfo{person}{Gunther Paul}.}
  \bibinfo{year}{2014}\natexlab{}.
\newblock \bibinfo{booktitle}{\emph{Huygens' principle and hyperbolic
  equations}}.
\newblock \bibinfo{publisher}{Academic Press}, \bibinfo{address}{Venice}.
\newblock


\bibitem[\protect\citeauthoryear{Paulin-Henriksson, Amara, Voigt, Refregier,
  and Bridle}{Paulin-Henriksson et~al\mbox{.}}{2008}]%
        {paulin2008point}
\bibfield{author}{\bibinfo{person}{S Paulin-Henriksson}, \bibinfo{person}{A
  Amara}, \bibinfo{person}{L Voigt}, \bibinfo{person}{A Refregier}, {and}
  \bibinfo{person}{SL Bridle}.} \bibinfo{year}{2008}\natexlab{}.
\newblock \showarticletitle{Point spread function calibration requirements for
  dark energy from cosmic shear}.
\newblock \bibinfo{journal}{\emph{Astronomy \& Astrophysics}}
  \bibinfo{volume}{484}, \bibinfo{number}{1} (\bibinfo{year}{2008}),
  \bibinfo{pages}{67--77}.
\newblock


\bibitem[\protect\citeauthoryear{Peng, Sun, Dun, Wetzstein, Heidrich, and
  Heide}{Peng et~al\mbox{.}}{2019}]%
        {peng2019learned}
\bibfield{author}{\bibinfo{person}{Yifan Peng}, \bibinfo{person}{Qilin Sun},
  \bibinfo{person}{Xiong Dun}, \bibinfo{person}{Gordon Wetzstein},
  \bibinfo{person}{Wolfgang Heidrich}, {and} \bibinfo{person}{Felix Heide}.}
  \bibinfo{year}{2019}\natexlab{}.
\newblock \showarticletitle{Learned large field-of-view imaging with thin-plate
  optics.}
\newblock \bibinfo{journal}{\emph{ACM Trans. Graph.}} \bibinfo{volume}{38},
  \bibinfo{number}{6} (\bibinfo{year}{2019}), \bibinfo{pages}{219--1}.
\newblock


\bibitem[\protect\citeauthoryear{Ratnasingam}{Ratnasingam}{2019}]%
        {ratnasingam2019deep}
\bibfield{author}{\bibinfo{person}{Sivalogeswaran Ratnasingam}.}
  \bibinfo{year}{2019}\natexlab{}.
\newblock \showarticletitle{Deep camera: A fully convolutional neural network
  for image signal processing}. In \bibinfo{booktitle}{\emph{Proceedings of the
  IEEE International Conference on Computer Vision Workshops}}.
  \bibinfo{publisher}{IEEE}, \bibinfo{address}{Venice}, \bibinfo{pages}{0--0}.
\newblock


\bibitem[\protect\citeauthoryear{Ren, Zhang, Wang, Hu, and Zuo}{Ren
  et~al\mbox{.}}{2020}]%
        {ren2020neural}
\bibfield{author}{\bibinfo{person}{Dongwei Ren}, \bibinfo{person}{Kai Zhang},
  \bibinfo{person}{Qilong Wang}, \bibinfo{person}{Qinghua Hu}, {and}
  \bibinfo{person}{Wangmeng Zuo}.} \bibinfo{year}{2020}\natexlab{}.
\newblock \showarticletitle{Neural blind deconvolution using deep priors}. In
  \bibinfo{booktitle}{\emph{Proceedings of the IEEE/CVF Conference on Computer
  Vision and Pattern Recognition}}. \bibinfo{publisher}{IEEE},
  \bibinfo{address}{Venice}, \bibinfo{pages}{3341--3350}.
\newblock


\bibitem[\protect\citeauthoryear{Ren, Zhang, Ma, Pan, Cao, Zuo, Liu, and
  Yang}{Ren et~al\mbox{.}}{2018}]%
        {ren2018deep}
\bibfield{author}{\bibinfo{person}{Wenqi Ren}, \bibinfo{person}{Jiawei Zhang},
  \bibinfo{person}{Lin Ma}, \bibinfo{person}{Jinshan Pan},
  \bibinfo{person}{Xiaochun Cao}, \bibinfo{person}{Wangmeng Zuo},
  \bibinfo{person}{Wei Liu}, {and} \bibinfo{person}{Ming-Hsuan Yang}.}
  \bibinfo{year}{2018}\natexlab{}.
\newblock \showarticletitle{Deep Non-Blind Deconvolution via Generalized
  Low-Rank Approximation}. In \bibinfo{booktitle}{\emph{Advances in Neural
  Information Processing Systems}},
  \bibfield{editor}{\bibinfo{person}{S.~Bengio}, \bibinfo{person}{H.~Wallach},
  \bibinfo{person}{H.~Larochelle}, \bibinfo{person}{K.~Grauman},
  \bibinfo{person}{N.~Cesa-Bianchi}, {and} \bibinfo{person}{R.~Garnett}}
  (Eds.), Vol.~\bibinfo{volume}{31}. \bibinfo{publisher}{Curran Associates,
  Inc.}, \bibinfo{address}{57 Morehouse Ln, Red Hook, NY 12571, USA}.
\newblock
\urldef\tempurl%
\url{https://proceedings.neurips.cc/paper/2018/file/0aa1883c6411f7873cb83dacb17b0afc-Paper.pdf}
\showURL{%
\tempurl}


\bibitem[\protect\citeauthoryear{Ronneberger, Fischer, and Brox}{Ronneberger
  et~al\mbox{.}}{2015}]%
        {ronneberger2015u}
\bibfield{author}{\bibinfo{person}{Olaf Ronneberger}, \bibinfo{person}{Philipp
  Fischer}, {and} \bibinfo{person}{Thomas Brox}.}
  \bibinfo{year}{2015}\natexlab{}.
\newblock \showarticletitle{U-net: Convolutional networks for biomedical image
  segmentation}. In \bibinfo{booktitle}{\emph{International Conference on
  Medical image computing and computer-assisted intervention}}. Springer,
  \bibinfo{publisher}{Springer}, \bibinfo{address}{Venice},
  \bibinfo{pages}{234--241}.
\newblock


\bibitem[\protect\citeauthoryear{Sajjadi, Scholkopf, and Hirsch}{Sajjadi
  et~al\mbox{.}}{2017}]%
        {sajjadi2017enhancenet}
\bibfield{author}{\bibinfo{person}{Mehdi~SM Sajjadi}, \bibinfo{person}{Bernhard
  Scholkopf}, {and} \bibinfo{person}{Michael Hirsch}.}
  \bibinfo{year}{2017}\natexlab{}.
\newblock \showarticletitle{Enhancenet: Single image super-resolution through
  automated texture synthesis}. In \bibinfo{booktitle}{\emph{Proceedings of the
  IEEE International Conference on Computer Vision}}.
  \bibinfo{publisher}{Springer}, \bibinfo{address}{Venice},
  \bibinfo{pages}{4491--4500}.
\newblock


\bibitem[\protect\citeauthoryear{Schroeder}{Schroeder}{1981}]%
        {schroeder1981modulation}
\bibfield{author}{\bibinfo{person}{Manfred~R Schroeder}.}
  \bibinfo{year}{1981}\natexlab{}.
\newblock \showarticletitle{Modulation transfer functions: Definition and
  measurement}.
\newblock \bibinfo{journal}{\emph{Acta Acustica united with Acustica}}
  \bibinfo{volume}{49}, \bibinfo{number}{3} (\bibinfo{year}{1981}),
  \bibinfo{pages}{179--182}.
\newblock


\bibitem[\protect\citeauthoryear{Schuler, Hirsch, Harmeling, and
  Schölkopf}{Schuler et~al\mbox{.}}{2016}]%
        {schuler2015learning}
\bibfield{author}{\bibinfo{person}{Christian~J. Schuler},
  \bibinfo{person}{Michael Hirsch}, \bibinfo{person}{Stefan Harmeling}, {and}
  \bibinfo{person}{Bernhard Schölkopf}.} \bibinfo{year}{2016}\natexlab{}.
\newblock \showarticletitle{Learning to Deblur}.
\newblock \bibinfo{journal}{\emph{IEEE Transactions on Pattern Analysis and
  Machine Intelligence}} \bibinfo{volume}{38}, \bibinfo{number}{7}
  (\bibinfo{year}{2016}), \bibinfo{pages}{1439--1451}.
\newblock
\urldef\tempurl%
\url{https://doi.org/10.1109/TPAMI.2015.2481418}
\showDOI{\tempurl}


\bibitem[\protect\citeauthoryear{Schwartzburg, Testuz, Tagliasacchi, and
  Pauly}{Schwartzburg et~al\mbox{.}}{2014}]%
        {schwartzburg2014high}
\bibfield{author}{\bibinfo{person}{Yuliy Schwartzburg}, \bibinfo{person}{Romain
  Testuz}, \bibinfo{person}{Andrea Tagliasacchi}, {and} \bibinfo{person}{Mark
  Pauly}.} \bibinfo{year}{2014}\natexlab{}.
\newblock \showarticletitle{High-Contrast Computational Caustic Design}.
\newblock \bibinfo{journal}{\emph{ACM Trans. Graph.}} \bibinfo{volume}{33},
  \bibinfo{number}{4}, Article \bibinfo{articleno}{74} (\bibinfo{date}{July}
  \bibinfo{year}{2014}), \bibinfo{numpages}{11}~pages.
\newblock
\showISSN{0730-0301}
\urldef\tempurl%
\url{https://doi.org/10.1145/2601097.2601200}
\showDOI{\tempurl}


\bibitem[\protect\citeauthoryear{Shi, Caballero, Husz{\'a}r, Totz, Aitken,
  Bishop, Rueckert, and Wang}{Shi et~al\mbox{.}}{2016}]%
        {shi2016real}
\bibfield{author}{\bibinfo{person}{Wenzhe Shi}, \bibinfo{person}{Jose
  Caballero}, \bibinfo{person}{Ferenc Husz{\'a}r}, \bibinfo{person}{Johannes
  Totz}, \bibinfo{person}{Andrew~P Aitken}, \bibinfo{person}{Rob Bishop},
  \bibinfo{person}{Daniel Rueckert}, {and} \bibinfo{person}{Zehan Wang}.}
  \bibinfo{year}{2016}\natexlab{}.
\newblock \showarticletitle{Real-time single image and video super-resolution
  using an efficient sub-pixel convolutional neural network}. In
  \bibinfo{booktitle}{\emph{Proceedings of the IEEE conference on computer
  vision and pattern recognition}}. \bibinfo{publisher}{Springer},
  \bibinfo{address}{Venice}, \bibinfo{pages}{1874--1883}.
\newblock


\bibitem[\protect\citeauthoryear{Shih, Guenter, and Joshi}{Shih
  et~al\mbox{.}}{2012}]%
        {shih2012image}
\bibfield{author}{\bibinfo{person}{Yichang Shih}, \bibinfo{person}{Brian
  Guenter}, {and} \bibinfo{person}{Neel Joshi}.}
  \bibinfo{year}{2012}\natexlab{}.
\newblock \showarticletitle{Image enhancement using calibrated lens
  simulations}. In \bibinfo{booktitle}{\emph{European Conference on Computer
  Vision}}. Springer, \bibinfo{publisher}{Springer}, \bibinfo{address}{Venice},
  \bibinfo{pages}{42--56}.
\newblock


\bibitem[\protect\citeauthoryear{Sliusarev}{Sliusarev}{1984}]%
        {sliusarev1984aberration}
\bibfield{author}{\bibinfo{person}{Georgii~Georgievich Sliusarev}.}
  \bibinfo{year}{1984}\natexlab{}.
\newblock \showarticletitle{Aberration and optical design theory}.
\newblock \bibinfo{journal}{\emph{ahl}} \bibinfo{volume}{17},
  \bibinfo{number}{10} (\bibinfo{year}{1984}), \bibinfo{pages}{1737--1754}.
\newblock


\bibitem[\protect\citeauthoryear{Smith}{Smith}{2005}]%
        {smith2005modern}
\bibfield{author}{\bibinfo{person}{Warren~J Smith}.}
  \bibinfo{year}{2005}\natexlab{}.
\newblock \bibinfo{booktitle}{\emph{Modern lens design}}.
\newblock \bibinfo{publisher}{American Mathematical Soc.},
  \bibinfo{address}{Boston}.
\newblock


\bibitem[\protect\citeauthoryear{Smith}{Smith}{2008}]%
        {smith2008modern}
\bibfield{author}{\bibinfo{person}{Warren~J Smith}.}
  \bibinfo{year}{2008}\natexlab{}.
\newblock \bibinfo{booktitle}{\emph{Modern optical engineering}}.
\newblock \bibinfo{publisher}{Tata McGraw-Hill Education},
  \bibinfo{address}{Boston}.
\newblock


\bibitem[\protect\citeauthoryear{Sun, Cao, Xu, and Ponce}{Sun
  et~al\mbox{.}}{2015}]%
        {sun2015learning}
\bibfield{author}{\bibinfo{person}{Jian Sun}, \bibinfo{person}{Wenfei Cao},
  \bibinfo{person}{Zongben Xu}, {and} \bibinfo{person}{Jean Ponce}.}
  \bibinfo{year}{2015}\natexlab{}.
\newblock \showarticletitle{Learning a convolutional neural network for
  non-uniform motion blur removal}. In \bibinfo{booktitle}{\emph{Proceedings of
  the IEEE Conference on Computer Vision and Pattern Recognition}}.
  \bibinfo{publisher}{IEEE}, \bibinfo{address}{Venice},
  \bibinfo{pages}{769--777}.
\newblock


\bibitem[\protect\citeauthoryear{Sun, Peng, and Heidrich}{Sun
  et~al\mbox{.}}{2017}]%
        {sun2017revisiting}
\bibfield{author}{\bibinfo{person}{Tiancheng Sun}, \bibinfo{person}{Yifan
  Peng}, {and} \bibinfo{person}{Wolfgang Heidrich}.}
  \bibinfo{year}{2017}\natexlab{}.
\newblock \showarticletitle{Revisiting cross-channel information transfer for
  chromatic aberration correction}. In \bibinfo{booktitle}{\emph{Proceedings of
  the IEEE International Conference on Computer Vision}}.
  \bibinfo{publisher}{IEEE}, \bibinfo{address}{Venice},
  \bibinfo{pages}{3248--3256}.
\newblock


\bibitem[\protect\citeauthoryear{Tao, Gao, Shen, Wang, and Jia}{Tao
  et~al\mbox{.}}{2018}]%
        {tao2018scale}
\bibfield{author}{\bibinfo{person}{Xin Tao}, \bibinfo{person}{Hongyun Gao},
  \bibinfo{person}{Xiaoyong Shen}, \bibinfo{person}{Jue Wang}, {and}
  \bibinfo{person}{Jiaya Jia}.} \bibinfo{year}{2018}\natexlab{}.
\newblock \showarticletitle{Scale-recurrent network for deep image deblurring}.
  In \bibinfo{booktitle}{\emph{Proceedings of the IEEE Conference on Computer
  Vision and Pattern Recognition}}. \bibinfo{publisher}{IEEE},
  \bibinfo{address}{Venice}, \bibinfo{pages}{8174--8182}.
\newblock


\bibitem[\protect\citeauthoryear{van~de Weijer, Gevers, and Gijsenij}{van~de
  Weijer et~al\mbox{.}}{2007}]%
        {van2007edge}
\bibfield{author}{\bibinfo{person}{Joost van~de Weijer}, \bibinfo{person}{Theo
  Gevers}, {and} \bibinfo{person}{Arjan Gijsenij}.}
  \bibinfo{year}{2007}\natexlab{}.
\newblock \showarticletitle{Edge-Based Color Constancy}.
\newblock \bibinfo{journal}{\emph{IEEE Transactions on Image Processing}}
  \bibinfo{volume}{16}, \bibinfo{number}{9} (\bibinfo{year}{2007}),
  \bibinfo{pages}{2207--2214}.
\newblock
\urldef\tempurl%
\url{https://doi.org/10.1109/TIP.2007.901808}
\showDOI{\tempurl}


\bibitem[\protect\citeauthoryear{Vedaldi and Fulkerson}{Vedaldi and
  Fulkerson}{2010}]%
        {vedaldi2010vlfeat}
\bibfield{author}{\bibinfo{person}{Andrea Vedaldi} {and} \bibinfo{person}{Brian
  Fulkerson}.} \bibinfo{year}{2010}\natexlab{}.
\newblock \showarticletitle{VLFeat: An open and portable library of computer
  vision algorithms}. In \bibinfo{booktitle}{\emph{Proceedings of the 18th ACM
  international conference on Multimedia}}. \bibinfo{publisher}{IEEE},
  \bibinfo{address}{Venice}, \bibinfo{pages}{1469--1472}.
\newblock


\bibitem[\protect\citeauthoryear{Wang, Chan, Yu, Dong, and Change~Loy}{Wang
  et~al\mbox{.}}{2019}]%
        {wang2019edvr}
\bibfield{author}{\bibinfo{person}{Xintao Wang}, \bibinfo{person}{Kelvin~CK
  Chan}, \bibinfo{person}{Ke Yu}, \bibinfo{person}{Chao Dong}, {and}
  \bibinfo{person}{Chen Change~Loy}.} \bibinfo{year}{2019}\natexlab{}.
\newblock \showarticletitle{Edvr: Video restoration with enhanced deformable
  convolutional networks}. In \bibinfo{booktitle}{\emph{Proceedings of the IEEE
  Conference on Computer Vision and Pattern Recognition Workshops}}.
  \bibinfo{publisher}{IEEE}, \bibinfo{address}{Venice}, \bibinfo{pages}{0--0}.
\newblock


\bibitem[\protect\citeauthoryear{Wang, Yu, Wu, Gu, Liu, Dong, Qiao, and
  Change~Loy}{Wang et~al\mbox{.}}{2018}]%
        {wang2018esrgan}
\bibfield{author}{\bibinfo{person}{Xintao Wang}, \bibinfo{person}{Ke Yu},
  \bibinfo{person}{Shixiang Wu}, \bibinfo{person}{Jinjin Gu},
  \bibinfo{person}{Yihao Liu}, \bibinfo{person}{Chao Dong}, \bibinfo{person}{Yu
  Qiao}, {and} \bibinfo{person}{Chen Change~Loy}.}
  \bibinfo{year}{2018}\natexlab{}.
\newblock \showarticletitle{Esrgan: Enhanced super-resolution generative
  adversarial networks}. In \bibinfo{booktitle}{\emph{Proceedings of the
  European Conference on Computer Vision (ECCV) Workshops}}.
  \bibinfo{publisher}{IEEE}, \bibinfo{address}{Venice}, \bibinfo{pages}{0--0}.
\newblock


\bibitem[\protect\citeauthoryear{Westphal, Rollins, Radhakrishnan, and
  Izatt}{Westphal et~al\mbox{.}}{2002}]%
        {westphal2002correction}
\bibfield{author}{\bibinfo{person}{Volker Westphal}, \bibinfo{person}{Andrew~M
  Rollins}, \bibinfo{person}{Sunita Radhakrishnan}, {and}
  \bibinfo{person}{Joseph~A Izatt}.} \bibinfo{year}{2002}\natexlab{}.
\newblock \showarticletitle{Correction of geometric and refractive image
  distortions in optical coherence tomography applying Fermat’s principle}.
\newblock \bibinfo{journal}{\emph{Optics Express}} \bibinfo{volume}{10},
  \bibinfo{number}{9} (\bibinfo{year}{2002}), \bibinfo{pages}{397--404}.
\newblock


\bibitem[\protect\citeauthoryear{Yan, Gong, Shi, Hengel, Shen, Reid, and
  Zhang}{Yan et~al\mbox{.}}{2019}]%
        {yan2019attention}
\bibfield{author}{\bibinfo{person}{Qingsen Yan}, \bibinfo{person}{Dong Gong},
  \bibinfo{person}{Qinfeng Shi}, \bibinfo{person}{Anton van~den Hengel},
  \bibinfo{person}{Chunhua Shen}, \bibinfo{person}{Ian Reid}, {and}
  \bibinfo{person}{Yanning Zhang}.} \bibinfo{year}{2019}\natexlab{}.
\newblock \showarticletitle{Attention-guided network for ghost-free high
  dynamic range imaging}. In \bibinfo{booktitle}{\emph{Proceedings of the IEEE
  Conference on Computer Vision and Pattern Recognition}}.
  \bibinfo{publisher}{IEEE}, \bibinfo{address}{Venice},
  \bibinfo{pages}{1751--1760}.
\newblock


\bibitem[\protect\citeauthoryear{Zhang, Zuo, Gu, and Zhang}{Zhang
  et~al\mbox{.}}{2017}]%
        {zhang2017learning}
\bibfield{author}{\bibinfo{person}{Kai Zhang}, \bibinfo{person}{Wangmeng Zuo},
  \bibinfo{person}{Shuhang Gu}, {and} \bibinfo{person}{Lei Zhang}.}
  \bibinfo{year}{2017}\natexlab{}.
\newblock \showarticletitle{Learning Deep CNN Denoiser Prior for Image
  Restoration}. In \bibinfo{booktitle}{\emph{IEEE Conference on Computer Vision
  and Pattern Recognition}}. \bibinfo{publisher}{IEEE},
  \bibinfo{address}{Venice}, \bibinfo{pages}{3929--3938}.
\newblock


\bibitem[\protect\citeauthoryear{Zhang, Chen, Ng, and Koltun}{Zhang
  et~al\mbox{.}}{2019}]%
        {zhang2019zoom}
\bibfield{author}{\bibinfo{person}{Xuaner Zhang}, \bibinfo{person}{Qifeng
  Chen}, \bibinfo{person}{Ren Ng}, {and} \bibinfo{person}{Vladlen Koltun}.}
  \bibinfo{year}{2019}\natexlab{}.
\newblock \showarticletitle{Zoom to learn, learn to zoom}. In
  \bibinfo{booktitle}{\emph{Proceedings of the IEEE Conference on Computer
  Vision and Pattern Recognition}}. \bibinfo{publisher}{IEEE},
  \bibinfo{address}{Venice}, \bibinfo{pages}{3762--3770}.
\newblock


\bibitem[\protect\citeauthoryear{Zhenggang}{Zhenggang}{2016}]%
        {zhenggang2016periscope}
\bibfield{author}{\bibinfo{person}{LI Zhenggang}.}
  \bibinfo{year}{2016}\natexlab{}.
\newblock \bibinfo{title}{Periscope lens and terminal device}.
\newblock
\newblock
\newblock
\shownote{US Patent 9,523,847.}


\bibitem[\protect\citeauthoryear{Zhu, Hu, Lin, and Dai}{Zhu
  et~al\mbox{.}}{2019}]%
        {zhu2019deformable}
\bibfield{author}{\bibinfo{person}{Xizhou Zhu}, \bibinfo{person}{Han Hu},
  \bibinfo{person}{Stephen Lin}, {and} \bibinfo{person}{Jifeng Dai}.}
  \bibinfo{year}{2019}\natexlab{}.
\newblock \showarticletitle{Deformable convnets v2: More deformable, better
  results}. In \bibinfo{booktitle}{\emph{Proceedings of the IEEE Conference on
  Computer Vision and Pattern Recognition}}. \bibinfo{publisher}{IEEE},
  \bibinfo{address}{Venice}, \bibinfo{pages}{9308--9316}.
\newblock


\end{thebibliography}

\end{document}